\documentclass[10pt,journal,compsoc]{IEEEtran}
\usepackage{bm}
\usepackage{color}
\usepackage{graphicx}
\usepackage{amsmath}
\usepackage{amssymb}
\usepackage{booktabs}
\usepackage{xcolor}
\usepackage{multirow}
\usepackage{diagbox}
\usepackage{makecell}

\newcommand{\red}[1] {\textcolor[rgb]{1.0,0.0,0.0}{{#1}}}
\newcommand{\blue}[1] {\textcolor[rgb]{0.0,0.0,1.0}{{#1}}}

\def\N{\mathbf{N}}
\def\I{\mathbf{I}}

\def\G{\mathbf{G}}
\def\g{\mathbf{g}}

\def\S{{\mathbf S}}

\def\n{{\mathbf n}}

\def\x{{\mathbf x}}
\def\s{{\mathbf s}}

\def\d{{\mathbf d}}
\def\v{{\mathbf v}}
\def\c{{\mathbf c}}

\def\c{{\mathbf c}}

\def\F{{\mathbf F}}

\ifCLASSOPTIONcompsoc

  \usepackage[nocompress]{cite}
\else
  \usepackage{cite}
\fi

\begin{document}

\title{Mesh Denoising Transformer}

\author{Wenbo Zhao,~\IEEEmembership{Member,~IEEE,}
        Xianming Liu,~\IEEEmembership{Member,~IEEE,}
        Deming Zhai,~\IEEEmembership{Member,~IEEE,}
        Junjun Jiang,~\IEEEmembership{Member,~IEEE,}
        and~Xiangyang Ji,~\IEEEmembership{Member,~IEEE}
\IEEEcompsocitemizethanks{\IEEEcompsocthanksitem W. Zhao, X. Liu, D. Zhai, J. Jiang are with the Faculty
of Computing, Harbin Institute of Technology, Harbin,
China, 150001.\protect\\
E-mail: \{csxm,zhaideming,jiangjujun\}@hit.edu.cn
\IEEEcompsocthanksitem X. Ji is with the Department of Automation, Tsinghua University, Beijing 100084, China.  E-mail: xyji@tsinghua.edu.cn.}}

\markboth{Journal of \LaTeX\ Class Files,~Vol.~14, No.~8, August~2015}%
{Shell \MakeLowercase{\textit{et al.}}: Bare Demo of IEEEtran.cls for Computer Society Journals}

\IEEEtitleabstractindextext{%
\begin{abstract}

Mesh denoising, aimed at removing noise from input meshes while preserving their feature structures, is a practical yet challenging task. Despite the remarkable progress in learning-based mesh denoising methodologies in recent years, their network designs often encounter two principal drawbacks: a dependence on single-modal geometric representations, which fall short in capturing the multifaceted attributes of meshes, and a lack of effective global feature aggregation, hindering their ability to fully understand the mesh's comprehensive structure. To tackle these issues, we propose \textit{SurfaceFormer}, a pioneering Transformer-based mesh denoising framework. Our first contribution is the development of a new representation known as  \textit{Local Surface Descriptor}, which is crafted by establishing polar systems on each mesh face, followed by sampling points from adjacent surfaces using geodesics. The normals of these points are organized into 2D patches, mimicking images to capture local geometric intricacies, whereas the poles and vertex coordinates are consolidated into a point cloud to embody spatial information. This advancement surmounts the hurdles posed by the irregular and non-Euclidean characteristics of mesh data, facilitating a smooth integration with Transformer architecture. Next, we propose a dual-stream structure consisting of a \textit{Geometric Encoder} branch and a \textit{Spatial Encoder} branch, which jointly encode local geometry details and spatial information to fully explore multimodal information for mesh denoising. A subsequent \textit{Denoising Transformer} module receives the multimodal information and achieves efficient global feature aggregation through self-attention operators. Our experimental evaluations demonstrate that this novel approach outperforms existing state-of-the-art methods in both objective and subjective assessments, marking a significant leap forward in the field of mesh denoising. 

\end{abstract}

\begin{IEEEkeywords}
mesh denoising, transformer, surface descriptor, multimodal.

\end{IEEEkeywords}}
\maketitle

\IEEEdisplaynontitleabstractindextext
\IEEEpeerreviewmaketitle

\section{Introduction}

\IEEEPARstart{M}{eshes} play an integral role in the domain of computer graphics and geometric processing, serving as the backbone for 3D shape representation, manipulation, and visualization. Their applications span a wide range of fields, including but not limited to 3D printing \cite{mao2018generating}, medical imaging \cite{wei2015morphology}, and augmented reality\cite{hamidian2010exploring}, enabling the detailed and flexible modeling of intricate geometries. The critical role of meshes in these applications underscores the importance of maintaining high-quality mesh data. However, meshes are prone to noise, which can stem from various sources such as the acquisition process (e.g., through 3D scanning) or subsequent processing stages, leading to a degradation in quality. This challenge has catalyzed significant research in the area of mesh denoising, which aims to eliminate noise while preserving vital geometric features like edges and corners.

Mesh denoising represents a complex and challenging ill-posed problem, aiming to smooth noisy surfaces while preserving the original features of the object and preventing unnatural geometric distortions. At the early stage, mesh denoising strategies were inspired by image denoising techniques, utilizing methods based on filtering~\cite{tomasi1998bilateral, li2014weighted} or optimization~\cite{xu2011image,elad2006image}. However, the unique attributes of mesh data, such as shape and topological relationships, make image-specific algorithms less effective for this task. As a result, these strategies often depend on low-dimensional geometric representations to simplify algorithm implementation and introduce additional prior knowledge about the shape or noise to aid in denoising. For instance, filtering-based approaches~\cite{zheng2011bilateral,zhang2015guided,GGNF} typically utilize face normals and center distances as geometric representations, leveraging these to determine filtering weights. Meanwhile, optimization-based approaches~\cite{he2013mesh,wu2015mesh,zhao2018robust} generally assume Gaussian-distributed noise and incorporate shape priors like piecewise-smooth constraints. These methods prove effective for CAD-like models, which are characterized by structured shapes and planar surfaces. However, they often struggle with complex or scanned models, leading to over-smoothing or the introduction of pseudo features due to the loss of mesh attributes and the limited generalization capability of the algorithms.

The advent of deep neural networks (DNNs) has significantly advanced data-driven mesh denoising~\cite{Wang-2016-SA,zhao2021normalnet, DBLP:journals/corr/abs-2108-05128}. DNNs are capable of directly mapping high-dimensional inputs to their ground truth, eliminating the need for complex inputs and additional prior knowledge. This has shifted the focus towards developing geometric representations that encapsulate more mesh attributes and network architectures with enhanced denoising capabilities. Early studies~\cite{Wang-2016-SA,wang2019data} employed filtering-based normal representation and utilized multilayer perceptron (MLP) networks for denoising. However, the limited expressive power of these representations and MLP's generalization capacity impacted denoising outcomes. Researchers have since explored more powerful architectures, divided into two categories: 1) some works attempted to create regular geometric representations, such as non-local normal matrices~\cite{li2020normalf} or voxel representations~\cite{zhao2021normalnet}, so as to leverage convolutional neural networks (CNNs) based networks; 2) others advocated for constructing graphs based on the topological structure of the mesh and preserving geometry information through node embedding, leading to the widespread adoption of graph neural networks (GNNs)~\cite{armando2020mesh, DBLP:journals/corr/abs-2108-05128, zhang2022geobi}. These approaches achieve superior denoising performance compared to traditional methods.

Despite these advancements, existing learning-based approaches still fall short in achieving satisfactory feature recovery results, which is attributed to two main factors. The first issue lies in the limitations of their network architectures, which are often confined to single-modal geometric representations. Such representations are insufficient for capturing the complex attributes of meshes, leading to a significant loss of attributes during the construction of representations. For instance, CNN-based methods, such as those discussed in~\cite{li2020normalf,zhao2021normalnet}, struggle to process the irregular spatial information inherent in meshes, whereas GNN-based methods~\cite{armando2020mesh, DBLP:journals/corr/abs-2108-05128, zhang2022geobi}, encounter difficulties in embedding detailed geometric information, resulting in the loss of geometric details. The second challenge is the networks' limited capability to capture global features. Typically, these networks attempt to accumulate global information through the stacking of convolutional layers or aggregations of graph nodes. However, the localized receptive fields of these layers are inadequate for managing long-range dependencies and understanding the overall structure of the mesh. This limitation hampers their ability to restore large-scale features that necessitate a comprehensive grasp of global structural information.

To overcome the above challenges, the exploration of advanced neural network architectures such as the Transformer~\cite{vaswani2017attention} for mesh denoising presents a promising avenue, due to the Transformer's ability to manage long-range dependencies and its flexibility in handling complex data structures. The Transformer's architecture, characterized by self-attention mechanisms, enables it to focus on relevant parts of the data regardless of their position, making it well-suited for tasks requiring an understanding of global context and structural relationships~\cite{wang2018non}. This feature is particularly relevant for mesh denoising, where capturing the global geometric structure and the intricate relationships between different parts of the mesh is crucial for effective denoising while preserving essential features.
Despite the potential advantages, the adoption of Transformer models for mesh denoising has been relatively slow, with a notable absence of research in this area. One of the primary reasons for this slow adoption is the challenge of data representation. The irregular and non-Euclidean characteristics of mesh data present substantial hurdles in formulating a representation that is compatible with Transformer models.  Unlike images or sequences, which have regular and structured data formats, meshes require specialized representations that can effectively capture their geometric and topological properties. 

Recent research in mesh-related tasks, such as mesh classification and segmentation~\cite{li2022laplacian}, mesh understanding~\cite{peng2023mwformer}, human pose and mesh reconstruction~\cite{lin2021mesh, lin2021end, lin2024mpt}, has explored to incorporate the Transformer architecture. These initiatives employed various encoders to handle multi-modal geometric representations and aggregate global information through the self-attention operator. Despite these advancements, current Transformer-based approaches in mesh processing often rely on existing representations like point clouds~\cite{li2022laplacian} or heat maps~\cite{lin2024mpt}, which may result in attribute loss. Furthermore, these networks are primarily designed for tasks of mesh understanding~\cite{li2022laplacian, peng2023mwformer} or 3D human pose and mesh reconstruction~\cite{lin2021mesh, lin2021end, lin2024mpt}, which are not directly applicable to mesh denoising without significant modifications.

In this study, we introduce a novel geometric representation called \textit{Local Surface Descriptors} (LSDs) to bridge this gap. LSDs are generated by setting up polar systems on each mesh face, from which points on adjacent surfaces are sampled using geodesic rays. The normals of these points are then arranged into 2D patches resembling images to capture local geometric details, while the poles are treated as a point cloud to encapsulate spatial information. This innovation overcomes the challenge posed by the irregular and non-Euclidean nature of mesh data, enabling seamless integration with Transformer models. We further present \textit{SurfaceFormer}, a powerful Transformer-based mesh denoising approach that employs a dual-stream structure to process multi-modal geometric representations and harnesses the Transformer architecture's superior global feature aggregation capabilities. Inspired by the success of Transformers in multi-modal tasks, \textit{SurfaceFormer} includes a Geometric Encoder branch and a Spatial Encoder branch to process local geometric and spatial information, respectively, converging in a Denoising Transformer module designed for effective feature aggregation. Our experimental evaluations demonstrate that \textit{SurfaceFormer} outperforms existing state-of-the-art mesh denoising methods, marking a significant leap forward in the field.

\begin{figure*}
\centering
\includegraphics[width=0.95\linewidth]{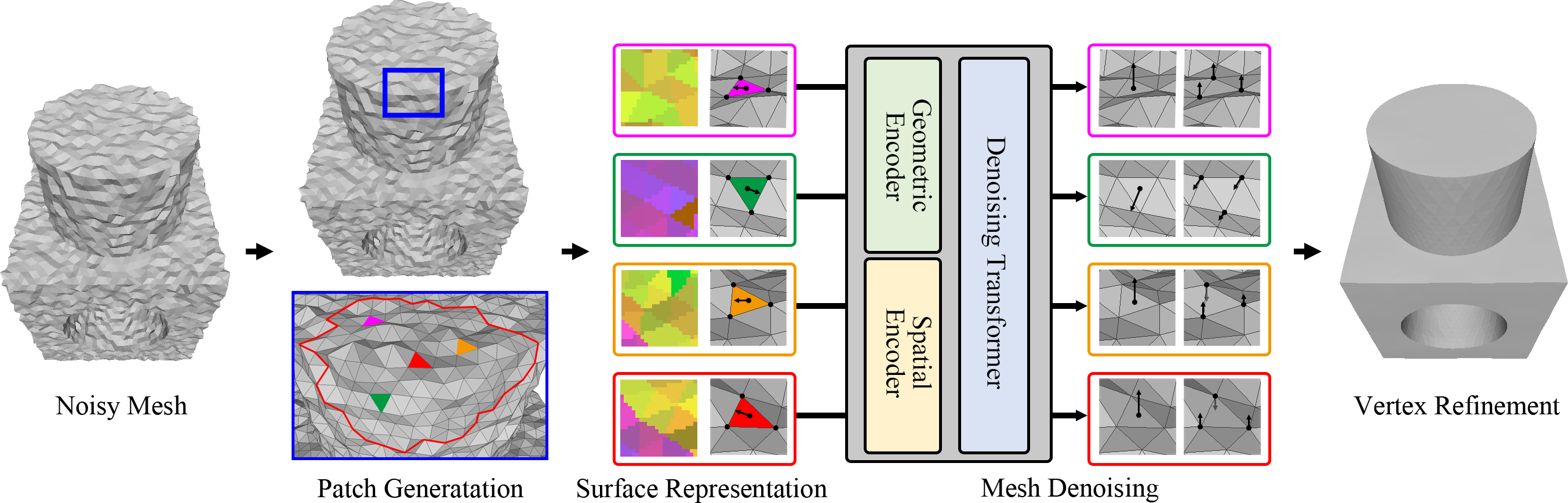}
    \caption{The overall framework of the proposed method, which includes four main steps: \textbf{Patch Generation}, \textbf{Surface Representation},  \textbf{Mesh Denoising} and \textbf{Vertex Refinement}. 
}
\label{framework}
\end{figure*}

The primary contributions of this work are highlighted as follows:
\begin{itemize}
\item  We introduce \textit{SurfaceFormer}, a pioneering Transformer-based mesh denoising framework that innovatively achieves multimodal geometric representation and global feature aggregation. This is, to the best of our knowledge, the first instance of applying Transformer architecture within the mesh denoising domain.

\item  We propose a novel representation called \textit{Local Surface Descriptors}, which efficiently preserves both the local geometry details and spatial information in a unified format, enabling smooth integration with Transformer architecture. 

\item  We offer extensive experimental results to demonstrate that \textit{SurfaceFormer} achieves substantial enhancements over current state-of-the-art methods in both subjective and objective evaluations, thanks to innovative network structures and geometric representations.

\end{itemize}

This paper significantly extends our preliminary work \cite{Zhao_Liu_Jiang_Zhao_Li_Ji_2022}, with notable advancements. Firstly, we refine the network architecture by integrating Transformers and introducing self-attention mechanisms. This development considerably boosts the network's ability to extract and synthesize features, addressing previously identified denoising limitations. Secondly, the representation design is optimized by incorporating multimodal geometric representations, which effectively retain both local geometric and spatial information. This approach successfully mitigates the feature loss issue seen with single-modal representations in our initial version. Furthermore, we perform an extensive experimental analysis against the latest state-of-the-art methods, showcasing the superior performance and effectiveness of our approach.


The remainder of this paper is structured as follows: Section 2 offers a concise overview of related works. The methodology behind \textit{SurfaceFormer} is detailed in Section 3. Experimental results are discussed in Section 4, and Section 5 provides the conclusion of the paper.

\section{Related Works}
\label{sec:related}
 


In this section, we review the related works from three aspects, including traditional approaches, data-driven approaches andd GNN-based approaches

\subsection{Traditional Approaches} 

Traditional approaches to mesh denoising have largely been developed from  image denoising techniques, including filtering and optimization approaches. Early technologies typically utilized a one-stage framework that directly applies these techniques to noisy vertex positions. For instance,  Fleishman \textit{et al.}~\cite{fleishman2003bilateral} and Jones \textit{et al.}~\cite{jones2003non} employed bilateral filter for feature-preserving denoising.  Clarenz \textit{et al.}~\cite{clarenz2000anisotropic,bajaj2003anisotropic, 10.1145/3596711.3596729} adapted the anisotropic diffusion filtering for mesh denoising. He \textit{et al.}~\cite{he2013mesh} introduced $L_0$ minimization for mesh denoising, leveraging differential edge operator and triangle shape regularization as optimization terms.  Wu \textit{et al.}~\cite{wu2015mesh} introduced $L_1$ minimization, incorporating vertex fidelity, total variation-based edge and triangle shape regularization. However, subsequent studies~\cite{zheng2011bilateral,zhang2015guided} indicated  that face normals offer a more accurate representation of local surface geometry than vertex positions. Consequently, the two-stage framework has gained popularity,  which firstly denoises the face normals and then update the vertex positions according to the denoised normals. For instance, Zheng \textit{et al.}~\cite{zheng2011bilateral} applied bilateral filter to face normals, determining filtering weights by center and normal distances. Zhang \textit{et al.}~\cite{zhang2015guided} introduced guided filter for mesh denoising, utilizing the average normal of the most consistent 1-ring sub-patch as guidance normals. Zhao \textit{et al.} ~\cite{zhao2019graph}proposed to use graph cuts to obtain feature-aware sub-patches for computing guidance normals. Recent efforts have sought to incorporate non-local information. This includes strategies such as adding non-local normals from similar patches during filtering~\cite{zhao2018multi} and organizing normals from similar patches into matrices, followed by low-rank recovery~\cite{wei2018mesh}.

Nevertheless, these methods generally require specific prior knowledge, such as the Gaussian noise distribution or piecewise-smooth shape prior. Consequently, these methods are only effective for CAD-like models that with simple structure, when facing complex or scanned models,  they fail to produce satisfactory results, often leading to over-smoothing or introducing pseudo-features.

\begin{figure*}
\centering
\includegraphics[width=0.95\linewidth]{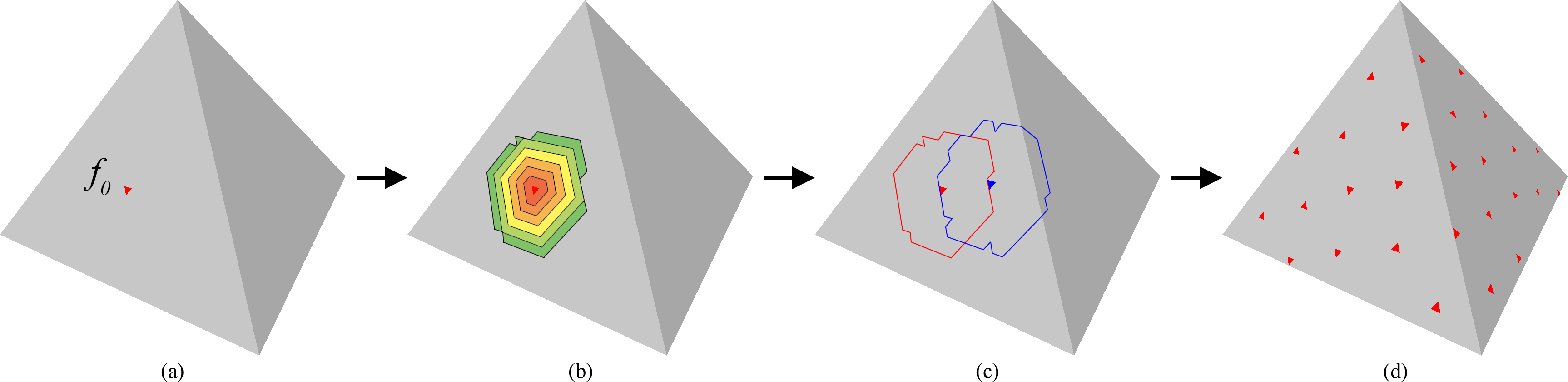}
    \caption{The patch generation process of \textit{Pyramid} when $T_f=240$. (a) $f_0$ is employed as the initial patch center, highlighted in red. (b) K-ring faces surrounding $f_0$ are incorporated into the patch until the number of faces reaches $T_f$. (c)  The nearest unvisited face to $f_0$ becomes the next center, highlighted in blue. Another patch is built around it. (d) This procedure continues until all the faces are visited, and the patch centers are highlighted in red.
}
\label{patch}
\end{figure*}

\vspace{-0.2cm}
\subsection{Data-driven approaches}
To address the aforementioned problems, researchers have embarked on exploring data-driven algorithms that directly learn the mapping from noisy data to ground truth. However, neural networks typically require regular input, and the irregular structure of mesh prevents the implement of neural networks. To tackle this problem, researchers have proposed a variety of geometric representations to convert irregular mesh information into regular structures. For instance, Wang \textit{et al.}~\cite{Wang-2016-SA} proposed to iteratively apply bilateral or guided filter on noisy meshes, the filtered normals are then organized as the filtered facet normal descriptor, and fed to MLP to obtain denoised normals. Wei \textit{et al.}~\cite{wei2019mesh} and Wang \textit{et al.}~\cite{wang2019data} improved this method by introducing generalized reverse descriptor and low-rank matrix recovery based preprocess. Nevertheless, the limited ability of descriptors to preserve features hinders the performance of these algorithms. To tackle this challenge, researchers have explored the construction of more efficient geometric representations.  Li \textit{et al.}~\cite{li2020dnf} represent local topological information as adjacency matrices, and use cascaded residual units to predict the denoised normals. Li \textit{et al.} ~\cite{li2020normalf} searched similar patches on the whole meshes to explore non-local information, and organized their normals to matrix, which allows the implement of CNN. Zhu \textit{et al.} ~\cite{zhu2021cascaded} built 2D height map to present the local geometric information. Zhao \textit{et al.} ~\cite{zhao2021normalnet} introduce voxelization to roughly represent the local shape information, and employ 3D Resnet for denoising.

These methods have made significant progress in preserving geometric information. However, the spatial information of mesh is difficult to be converted into a regular form, and is therefore usually discarded, which reduces the ability of the algorithm to distinguish feature structures.

\subsection{GNN based Approaches}
Since meshes inherently possess a graph structure defined by vertex and face connectivity, prompting researchers to leverage graph neural networks (GNNs) for enhancing spatial information preservation. Armando \textit{et al.}~\cite{armando2020mesh} pioneered GNN-based method for face normal denoising by treating each face as a node and building edges between adjacent faces. Shen \textit{et al.}~\cite{DBLP:journals/corr/abs-2108-05128} employed static and dynamic graph convolutions to connect non-adjacent graph nodes. Zhang \textit{et al.}~\cite{zhang2022geobi} proposed building graphs on both faces and vertices, enabling simultaneous denoising in normal and vertex domains, followed by a vertex refinement strategy to align denoised normals and vertex coordinates. Hattori \textit{et al.}~\cite{hattori2022learning} also constructed two graphs and introduced self priors, such as smoothness error and normal consistency, to achieve self-supervised mesh denoising. Recently, Tang \textit{et al.} proposed DOGNET~\cite{tang2023feature} consisting of cascaded networks that conduct localized graph neural network operations to aggregate features from neighborhoods, and iteratively propagate information across the mesh. Zhao \textit{et al.}~\cite{zhao2023multi, zhao2024curvature} further introduced curvature domain denoising to improve the feature preserving denoising performance. Zhou \textit{et al.}~\cite{zhou2024resgem} proposed predicting vertex offsets instead of directly predicting positions, facilitating network convergence and yielding improved results.

While GNNs offer significant advantages over traditional neural networks in preserving spatial information, they often lack the design of geometric attribute embeddings, resulting in the loss of local geometric information and thereby limiting method performance.





\section{Methodology}
\label{sec:method}



\subsection{Overview}

Define $ \mathcal{X} = \left\{\mathcal{V}, \mathcal{F} \right\}$  as the input noisy mesh, where $\mathcal{V}=\{\v_i\}$ and $\mathcal{F}=\{f_i\}$ denote the sets of vertices and faces, respectively. Our objective is to preform denoising in both normal and vertex domains,  followed by aligning the denoised vertex coordinates and face normals to generate the denoised mesh $\widehat{\mathcal{X}}$. As illustrated in Fig. \ref{framework}, the proposed \textit{SurfaceFormer} encompasses four principal phases:
    

\begin{itemize}
    \item \textbf{Patch Generation.} This phase involves the segmentation of the input noise mesh $\mathcal{X}$ into equally-sized patches. By dividing the mesh into smaller, manageable patches, it allows for more focused and efficient processing of local geometries. The segmentation is designed to ensure that each patch contains a portion of the mesh that is sufficiently representative of the local structure, allowing for accurate local surface reconstruction in later phases.
    \item \textbf{Surface Representation.} In this phase, each segmented patch is encoded using the Local Surface Descriptor (LSD), which captures the mesh's local geometric details as well as spatial information. The geometric details are represented in the form of image-like 2D matrices, while the spatial information is captured as a point cloud. This dual representation is critical for preserving the intricate details of the mesh's surface, ensuring that the model has access to comprehensive information about the local geometry of each patch.
    \item \textbf{Mesh Denoising.} Upon representing the surface of each patch, the \textit{SurfaceFormer}  processes the LSD to mesh denoising. The process involves two main steps: the 2D matrices are fed into a Geometric encoder, and the point cloud into a Spatial encoder. These encoders extract and refine features from their respective inputs, which are then fused together. The fused features undergo further refinement in the Denoising Transformer, which accurately reconstructs the surface normals and vertex coordinates.
    \item \textbf{Vertex Refinement.} While denoised meshes can be obtained by using only denoised surface normals or vertex coordinates, this may introduce artificial effects to mesh and reduce surface consistency, respectively. Therefore, an iterative refinement strategy is employed to align denoised vertex coordinates with face normals to further refine the denoising results and output the final denoised meshes.

    
\end{itemize} 

By addressing both geometric and spatial aspects of the mesh and leveraging advanced encoding and transformation techniques, \textit{SurfaceFormer} promises enhanced accuracy and detail preservation in the denoised mesh output. In the following, we will introduce the four phases in details.

\subsection{Patch Generation} 


Given the extensive variety of mesh sizes, which can encompass up to millions of faces, the task of directly denoising an entire mesh proves to be too demanding for standard GPUs due to the substantial computational resources required. To circumvent this limitation, we employ a segmentation strategy inspired by Transformer-based methods for processing lengthy inputs~\cite{xie2020unsupervised, joshi2019bert}. This strategy involves dividing noisy meshes into overlapping patches. As demonstrated in Fig. \ref{patch}, we commence by selecting a face to serve as the center of a patch. For the sake of simplicity, the first face $f_0$ is chosen as this center. Following this, we progressively incorporate the $k$-ring neighbors of $f_0$ into the patch, continuing this process until the number of faces in the patch reaches a predetermined limit $T_f$. Subsequently, we identify the closest unvisited face to $f_0$ to serve as the center for the next patch, and this process is repeated, allowing us to construct patches until every face has been accounted for.

An important consideration of this segmentation approach is that it results in each face undergoing the denoising process multiple times, which in turn produces several denoised normals and vertex coordinates for each face. The approach to resolving this issue is further elaborated in the \textit{Vertex Refinement} section.


\subsection{Surface Representation}
We then delve into the detailed process of generating the LSD. Given a patch $P=\left\{f_i\right\}$, the goal of LSD is to convert the mesh attributes of each face in $P$ into two components with different modal: an image-like matrix $\g$ that captures local geometric details, and a featured point $\s$ that represents the spatial information. Specifically, for a face $f_i$, we uniformly sample a set of points $\left\{p_{\left(i,j\right)}\right\}, i,j\in\left(-T_s,T_s\right]$ around a chosen pole, where $T_s$ is a parameter determining the number of sampling points. The normals of these sampled points and the pole are then compiled into the LSD $\left(\g_i,\s_i\right)$ of $f_i$. Finally, these LSDs are aggregated into a matrix sequence $\G=\left\{\g_i\right\}$ and a point cloud $\S=\left\{\s_i\right\}$, which serve as the input to the denoising network.

\begin{figure}
\centering
\includegraphics[width=0.95\linewidth]{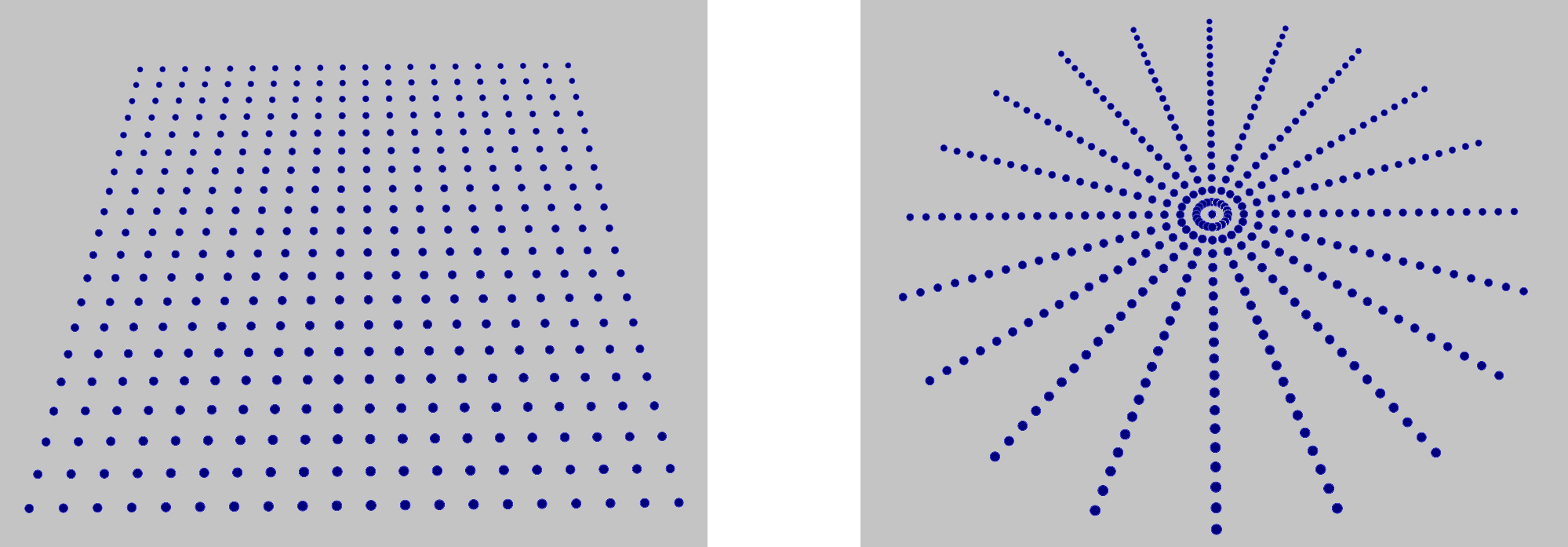}
{\hspace{0cm}(a)\hspace{4.5cm}(b)}
\caption{Sample $20\times20$ points (a) with uniformly distributed Cartesian coordinates, in which the sampling density is consistent. (b) with uniformly distributed polar coordinates, in which the sampling density tends to be sparser with increased radius.}
\label{fig_equidistantly}
\end{figure}

\subsubsection{Generation of LSD}

Addressing the complex challenge of achieving uniform sampling on the curved surfaces of meshes is pivotal for generating LSD. While uniform sampling on planar surfaces can be straightforwardly achieved by setting up a 2D Cartesian coordinate system and methodically choosing a set of Cartesian coordinates, as depicted in Fig.\ref{fig_equidistantly}-(a), this approach falls short when applied to curved surfaces.

\begin{figure}
\centering
\includegraphics[width=0.95\linewidth]{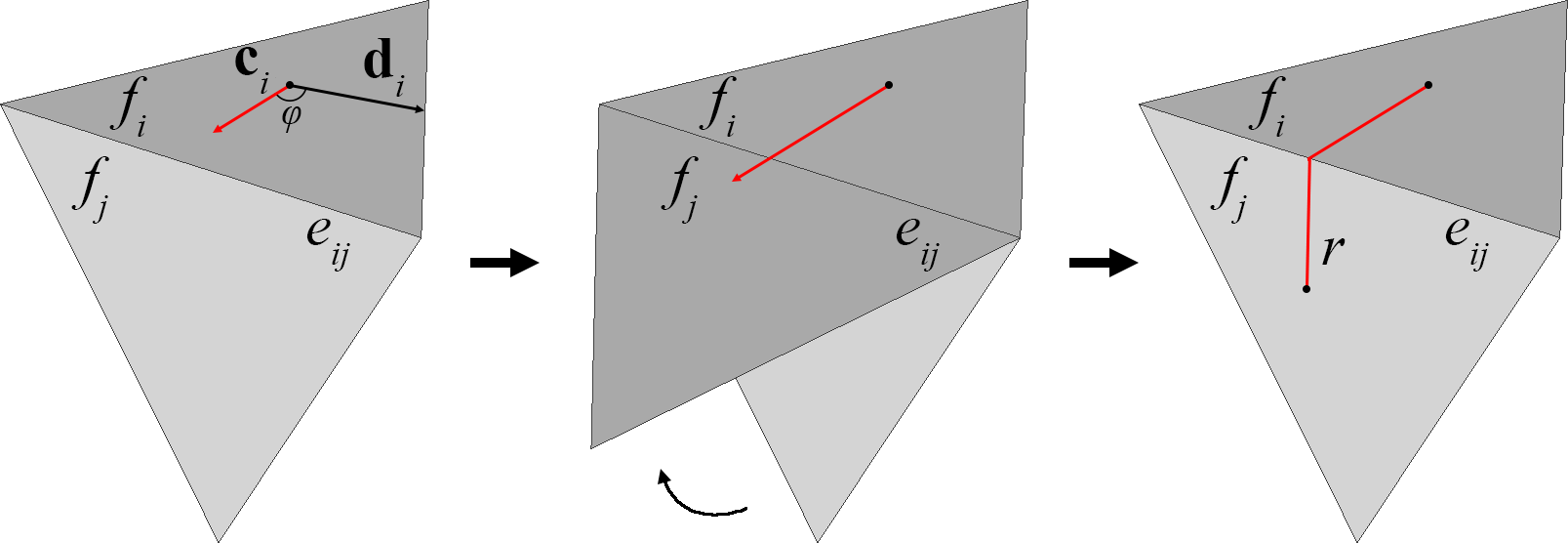}
{\hspace{0cm}(a)\hspace{2.7cm}(b)\hspace{2.7cm}(c)}
\caption{An example of sampling by shooting geodesic. The polar coordinate system is built on $f_i$, $\c_i$ and $\d_i$ are the polar and axis, respectively. (a) The geodesics starts from $\c_i$ at angle $\varphi$. (b) When the geodesics reaches edge $e_{ij}$, $f_j$ is rotated to the same plane as $f_i$ around $e_{ij}$ to allow the propagation. (c) The geodesics stops propagation when the length is equal to $r$, and the end point is regarded as the sampling point.}
\label{fig_propagation}
\end{figure}

Previous studies, such as \cite{KK2012}, introduced the use of a polar coordinate system $\left\{p_s\right\}=\left\{r,\varphi\right\}$ on mesh surfaces, where $r$ represents the radial distance and $\varphi$ the angular direction. In this method, geodesics are projected from a central point to divide the local surface into several sectors, gathering data within each sector to create a shape context descriptor. Drawing inspiration from this technique, we devise a sampling methodology that projects geodesics at specific polar coordinates and utilizes the endpoints as the sampling points. As illustrated in Fig.~\ref{fig_propagation}, considering a current face $f_i$, its neighboring face $f_j$, and their connecting edge $e_{ij}$, we use the center $\c_i$ of $f_i$ as the pole and a ray $\d_i$ from $\c_i$ to the midpoint of one of $f_i$'s edges as the axis. A geodesic is initiated from $\c_i$ at a specific angle $\varphi$ and, upon reaching $e_{ij}$, the adjoining face $f_j$ is aligned onto the same plane as $f_i$ by rotation around $e_{ij}$, facilitating the geodesic's continuation on this plane. This process repeats until the geodesic attains a total length of $r$, with the endpoint serving as the sampling point.

\begin{figure*}
\centering
\includegraphics[width=0.95\linewidth]{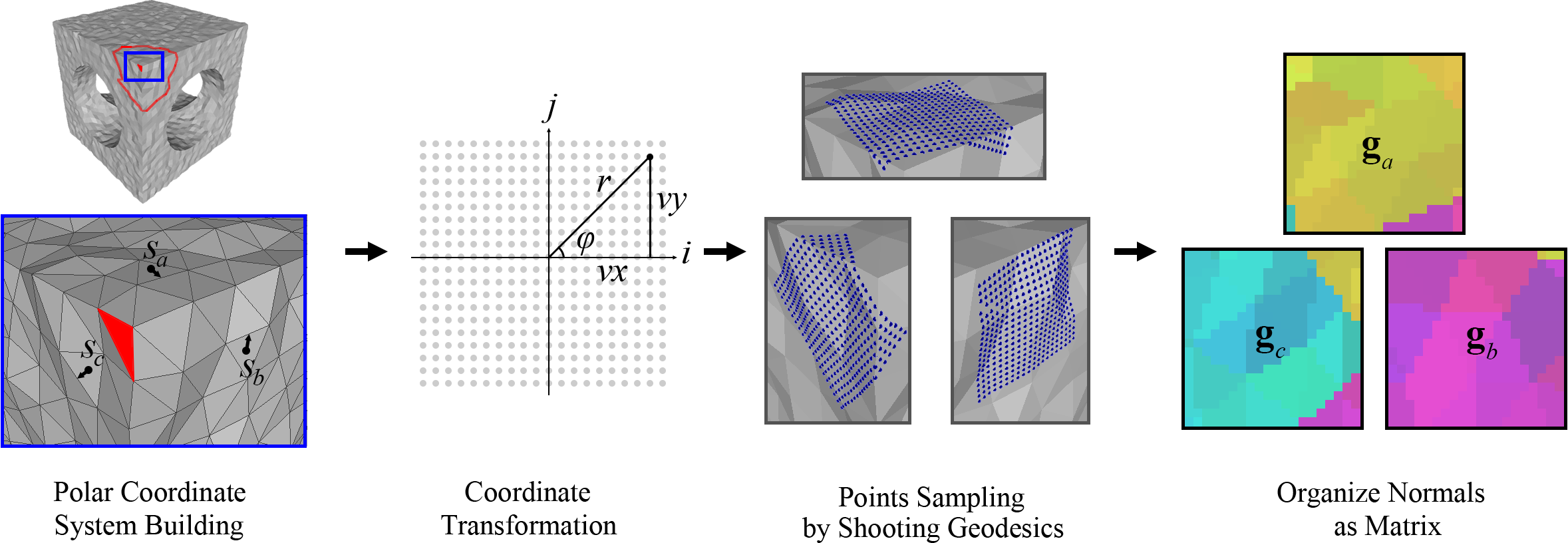}
    \caption{An illustration of the LSD generation with $T_s=10$.
}
\label{LSD}
\end{figure*}

Nevertheless, achieving uniform sampling remains a challenging task due to the fact that the sampling density gradually decreases with increasing $r$, as shown in Fig.~\ref{fig_equidistantly}-(b). To overcome this problem, we propose a coordinate transformation strategy that leverages the benefits of both Cartesian and polar coordinate systems. Specifically, our goal is to generate a polar coordinate set $\left\{\left(r_{\left(i,j\right)}, \varphi_{\left(i,j\right)} \right)\right\}, i,j\in\left(-T_s,T_s\right]$ using the coordinate transformation strategy. Initially, a virtual Cartesian system $\left\{c_s\right\}=\left\{vx,vy\right\}$ is established. A virtual coordinate set $\left\{\left(vx_i,vy_j \right)\right\}$ with square distribution is then generated in the system. To adapt the distance between sampling points to the mesh scale, the virtual Cartesian coordinates are calculated by:
\begin{equation}
\left(vx_i,vy_j\right)=\left(\frac{d_a}{p_s}i,\frac{d_a}{p_s}j \right)
\end{equation}
where $d_a$ represents the average Euclidean distance between adjacent face centers in the noisy mesh, serving as a scale estimate; ${p_s}$ is the parameter controlling sampling precision. Subsequently, the virtual Cartesian coordinates $\left(vx_i,vy_j\right)$ are transformed into polar coordinates $\left(r_{\left(i,j\right)}, \varphi_{\left(i,j\right)} \right)$:
\begin{equation}
\begin{array}{*{20}{c}}
  {r_{\left(i,j\right)} = \sqrt {{{\left( vx_i \right)}^2} + {{\left( vy_j \right)}^2}} },
  {\varphi_{\left(i,j\right)} = \arctan \frac{vy_j}{vx_i}} 
\end{array}
\end{equation}

After obtain the polar coordinate set, we adopt the sampling strategy to obtain regular sampling points, and their normals $\left\{\n_{\left(i,j \right)}\right\}$ of the sampling points are arranged into a matrix as $\g_i$. We also preserve the pole and the three vertex coordinates of $f_i$ as the spatial information $\s_i=\left[\c_i,\d_i,\v_i^1,\v_i^2,\v_i^3\right]$, where $\v_i^1,\v_i^2,\v_i^3$ denotes the vertices of $f_i$. Finally, we consolidate the LSDs of all the faces in $P$ to a matrix sequence $\G \in \mathbb{R} ^ {T_f \times \left( \left(2T_s\right)^2 \times 3\right)}$ and a point cloud $\S \in \mathbb{R} ^ {T_f \times 15}$. 

For better understanding, we present the LSD generation of three different faces in Fig.~\ref{LSD}, which presents the LSDs of three distinct faces. It reveals that the proposed strategy ensures that the sampling points are uniformly distributed on the surface of the object, and the normals of sampling points can accurately capture the feature structures around each face.

\subsubsection{Normalization}
We further implement a normalization strategy to ensure that the LSDs possess scale and rotation invariance. Specifically, we translates the patch center to the origin, and scale the patch to fit within [-1, 1]. Then we calculate the average normal ${\overline \n}$ of the faces in $P$, and utilize Rodrigues' rotation formula~\cite{liang2018efficient}  to derive the rotation matrix that aligns ${\overline \n}$ with a predetermined direction $\n_t$:
\begin{equation}
\begin{array}{*{20}{c}}
  {\theta = \angle (\overline \n, \n_t)},
  {\mathcal{R} ={\I}+\left(sin\theta\right){\N}+\left(1-cos\theta\right){\N^2}} 
\end{array}
\end{equation}
where $\N$ is the skew-symmetric cross-product matrix of ${\overline\n} \times {\n_t}$. 

The translation and rotation are then applied on the LSDs to get normalized $\tilde {\G}$ and $\tilde {\S}$:
\begin{equation}
\begin{array}{*{22}{c}}
  {\tilde \n_{\left(i,j \right)}={\mathcal{R}}\n_{\left(i,j \right)}},
  { \tilde {\d_i} ={\mathcal{R}}\d_i},\\
  { \tilde {\c_i}=\frac{{\mathcal{R}} \left(\c_i-\c_0\right)}{M_v}},
    { \tilde {\v_i}=\frac{{\mathcal{R}} \left(\v_i-\c_0\right)}{M_v}} 
\end{array}
\end{equation}
where $\c_0$ denotes the center of the patch center face, $M_v$ denotes the the maximum absolute value of all the coordinate components.

It is critical to note that the denoised normals and vertices derived from the normalized LSDs are also normalized. Therefore, we also calculate the inverse rotation matrix $\mathcal{R}^{-1}$ and keep $\c_0, M_v$ to facilitate the reversion of results after denoising.

\begin{figure*}
\centering
\includegraphics[width=0.95\linewidth]{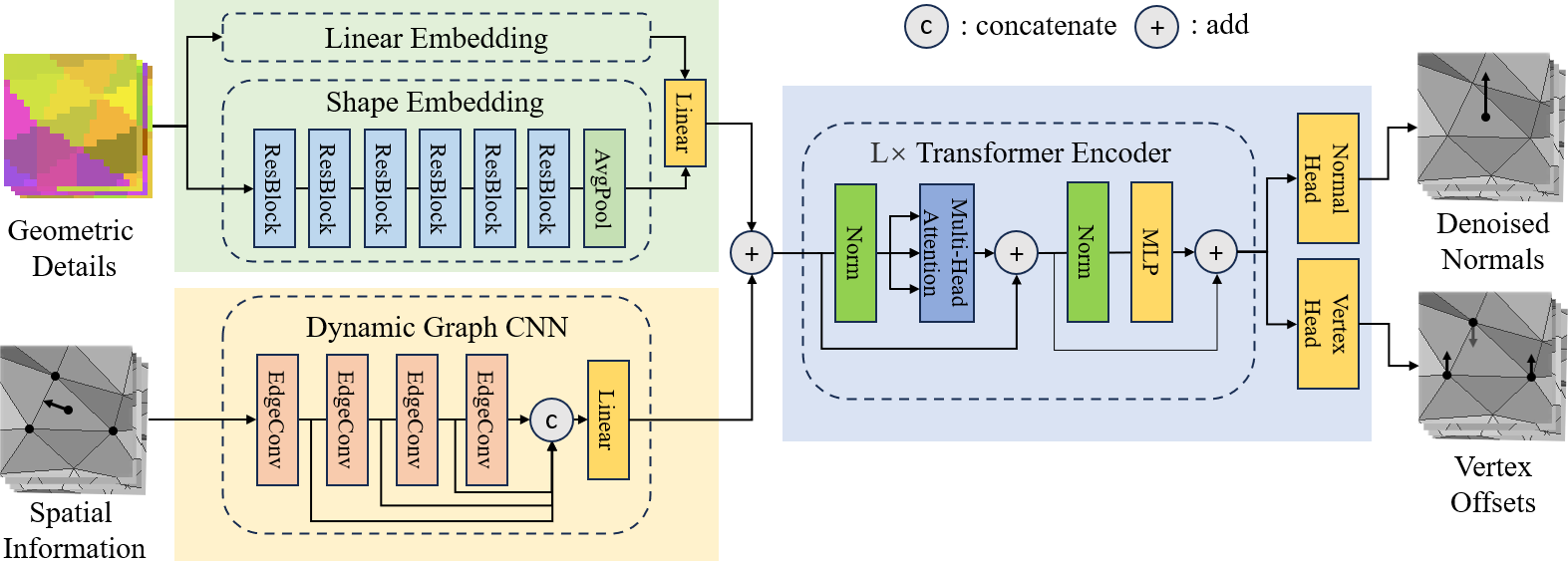}
    \caption{ The network architecture of \textit{SurfaceFormer}.
}
\label{network}
\end{figure*}
\subsection{Mesh Denoising}
The proposed LSD preserves both geometric and spatial information, which allows us to follow the multi-domain mesh denoising approaches~\cite{zhang2022geobi, zhao2024curvature, zhou2024resgem}, achieving normal and vertex denoising simultaneously. Specifically, the normalized LSDs are fed to the proposed \textit{SurfaceFormer} to predict the denoised face normals $\left\{\widehat\n_i\right\}$ and vertex coordinate offsets $\left\{\triangle \widehat\v_i\right\}$, which features a dual-stream structure, comprising a Geometric Encoder branch and a Spatial Encoder branch, culminating in a Denoising Transformer. As illustrated in Fig.~\ref{network}, the Geometric Encoder and Spatial Encoder collaboratively process $\tilde {\G}$ and $\tilde {\S}$, respectively, projecting them into $\F_g$ and $\F_s$ as geometric and spatial features. These features are then fed into the Denoising Transformer to achieve global feature aggregation and denoising. The functionality and details of these three key components will be elaborated upon in the subsequent subsections.

\subsubsection{Geometric Encoder}
Leveraging the similarity between the matrix sequence $\tilde {\G}$ and image patches in Visual Transformer (ViT)~\cite{dosovitskiy2020image}, employing the linear embedding from ViT has demonstrated effective denoising outcomes. However, it is crucial to recognize that $\tilde {\G}$ embodies the mesh's piecewise-smooth nature, in which the edges of faces play an important role in reflecting the shape information and feature structures. Linear embedding could potentially reduce the correlation among edges, resulting in the loss of feature details. To address this issue, as highlighted in the green box in Fig.~\ref{network}, the proposed Geometric Encoder incorporates both linear embedding and shape embedding for capturing normal and edge information, respectively. Given the input $\tilde {\G}$, the output $\F_g$ is generated as below:
\begin{equation}
{\F_g} = {L_m}\left(  \left[ {E_L}\left( \tilde {\G} \right), {E_C}\left( \tilde {\G} \right) \right] \right), {\F_g}  \in  \mathbb{R} ^ {T_f \times D}
\end{equation}
where ${E_L}$ represents the linear embedding from ViT~\cite{dosovitskiy2020image}, $E_{C}$ represents the shape embedding, which is a lightweight ResNet~\cite{7780459} that consist of 6 ResNet Blocks and an average pooling, each block featuring two $3\times3$ convolution layers with residual connections linking the input and output of each block, the channel number of all the convolution layers is 32. $L_m$ denotes a linear projection that fuses the outcomes of both embeddings, producing a $D$-dimensional geometric feature $\F_g$.
The Spatial Encoder serves a similar purpose to positional Embedding in Transformer architecture, equipping the network with the capability to recognize and leverage positional correlations. However, due to the variable nature of positional relationships in point clouds, employing a pre-trained weight matrix, as done in image or natural language processing domains~\cite{zhao2020exploring, vaswani2017attention}, is unfeasible. To address this challenge, we utilize the Dynamic Graph CNN~\cite{wang2019dynamic} as a trainable Spatial Encoder $E_{S}$  to capture local geometric details. Following the configuration in~\cite{wang2019dynamic}, our Spatial Encoder includes four EdgeConv operations designed to extract features across multiple scales, whose output dimension are 64, 64, 128 and 256, respectively. These extracted features are subsequently concatenated and processed through a linear layer to produce the spatial feature $\F_s$, defined as follows:

\begin{equation}
{\F_g} = {E_S}\left( \tilde {\S} \right), {\F_g} \in \mathbb{R} ^ {T_f \times D} 
\end{equation}
For the ease of feature fusion later in the process, we ensure that the dimensions of $\F_s$ match those of $\F_g$.

\subsubsection{Denoising Transformer}
The initial step involves combining the geometric and spatial features to obtain the fused feature. We adopt the design of ViT~\cite{dosovitskiy2020image} that achieves fusion through addition:
\begin{equation}
\F_0 = \F_g +  \F_s, {\F_0} \in \mathbb{R} ^ {T_f \times D} 
\end{equation}

Subsequently, we utilize $L\times$ Transformer Encoders to facilitate global feature aggregation. These encoders comprise multi-head self-attention ($\mathbb{MSA}$) and multilayer perceptrons ($\mathbb{MLP}$), both augmented with Layer Normalization ($\mathbb{LN}$) and residual connections. All the encoders maintains a constant latent feature with size $D$ and employ $N_h$ attention heads. The feature aggregation process is defined as follows:

\begin{equation}
{\F_i^{'}} = \mathbb{MSA}\left( {\mathbb{LN}\left( {{\F_{i - 1}}} \right)} \right) + {\F_{i - 1}}, i \in [1,L]
\end{equation}
\begin{equation}
{\F_i} = \mathbb{MLP}\left( {\mathbb{LN}\left( {\F_i^{'}} \right)} \right) + {\F_i^{'}}, i \in [1,L]
\end{equation}

The output $\F_{L}$ is then directed to the normal head $H_n$ and vertex head $H_v$ to produce the normalized denoised normal $\left\{\widehat\n_i\right\}$ and three vertex coordinate offsets  $\left\{\triangle \widehat\v_i\right\}$ of each face. For vertex denoising, we apply a linear layer with $\mathbb{LN}$ to yield 9-dimensional output values, formulated as:

\begin{equation}
\left\{ \triangle \tilde\v_i\right\}= H_v\left( {\mathbb{LN}\left( {{\F_{L}}} \right)} \right) 
\end{equation}
It is important to note that each vertex may belong to multiple faces, which also yields duplicate denoising results and will be resolved in \textit{Vertex Refinement} section.

For normal denoising, given that $\left\{\widehat\n_i\right\}$ are expected to be unit vectors, we apply a linear layer with $\mathbb{LN}$ and normalization ($\mathbb{NL}$) to yield 3-dimensional output values  $H_n$, formulated as:

\begin{equation}
\left\{ \tilde\n_i\right\}= \mathbb{NL}  {\left( H_n\left( {\mathbb{LN}\left( {{\F_{L}}} \right)} \right) \right)} 
\end{equation}

Finally, we apply the inverse rotation and transformation to recover the denoised normals and vertices:
\begin{equation}
\left\{ \widehat \n_i\right\}= {\mathcal{R}^{-1}} \left\{ \tilde\n_i\right\}, \left\{ \widehat \v_i\right\}= M_v {\mathcal{R}^{-1}} \left( \left\{ \v_i +\triangle \widehat\v_i \right\} +\c_0\right)
\end{equation}




\subsection{Vertex Refinement}
The proposed \textit{Patch Generation} and \textit{Mesh Denoising} strategy will produce multiple denoised normals and vertex coordinates for each face. To address this problem, we straightforwardly employ the average denoised normals $\overline {\n_i}$ and $\overline {\v_i}$ as the denoised results.

Upon obtaining the denoised normals and vertex coordinates, two common methods can be employed to obtain the final denoised mesh: updating vertex coordinates based on denoised normals~\cite{zhang2015guided} or directly utilizing denoised vertex coordinates~\cite{fleishman2003bilateral}. However, since existing vertex updating methods [9] fail to capture the underlying patterns of vertex distribution, the first method may introduce artificial effects to the denoised mesh. The second method can avoid this issue, but the denoised vertex coordinates demonstrate less local consistency than surface normals, resulting in the inability to produce smooth denoising results. To address this challenge, we employ the vertex refinement scheme~\cite{zhang2022geobi} which aligns the denoised vertex coordinates with face normals, thereby resolving the aforementioned problems:

\begin{equation}
{\overline {\v_i}}^{'} = {\overline {\v_i}} + \frac{1}{{\left| {\theta_{\v}} \right|}}\sum\limits_{ {f_i\in\theta_{\v}}} {\overline \n _i} \left(\overline \n_i \cdot\left({\c_i} - \overline {\v_i} \right) \right)
\end{equation}

where ${ {\theta_{\v}} }$ denotes the set of faces that include vertex ${\v}$. This refinement process is repeated for $N_v$ times and outputs the final denoised mesh $\widehat{\mathcal{X}}$.

\section{Network Training}
\label{Training}
The training set of \textit{SurfaceFormer} is constructed from a set of noisy meshes $\left\{{\mathcal{X}}\right\}$ and their corresponding noise-free meshes $\left\{{\mathcal{X^*}}\right\}$. Specifically, we build a patch around each face in $\left\{{\mathcal{X}}\right\}$, and implement \textit{Surface Representation} to obtain the normalized LSDs that serve as the input of network. Then we compute the corresponding normalized noise-free normals ${\tilde \n^*}$ and the offset ${\tilde \v^*}-\tilde{\v}$ from $\left\{{\mathcal{X^*}}\right\}$ as the ground truth. To prevent overfitting, we apply random rotations and jittering to the LSDs and ground truth. 

We train \textit{SurfaceFormer} on a server equipped with two RTX3090 GPUs and an AMD Ryzen 5900X CPU for $10^{5}$ iteration, optimized with the Adam algorithm ($\beta_1=0.9$, $\beta_2=0.999$, learning rate = 0.0001, batch size = 80). The loss function covers both the normal and vertex offset regressions, which are measured by the $L_1$ distance between outputs and ground truth:

\begin{equation}
L = \sum\limits_{{f_i} \in P} {{\left\|  \tilde {\n_i}^* - \tilde{\n_i} \right\|}_1}  + \alpha  \sum\limits_{{f_i} \in P} {{\left\|  \left(\tilde {\v_i}^* - \tilde{\v_i} \right) - \triangle \tilde\v_i  \right\|}_1}
\end{equation}
where $\alpha$ is the weight parameter to balance the two loss terms, we find that setting $\alpha=1$ can produce satisfactory denoising results.

\section{Experiments}
\label{sec:experiments}

This section presents the outcomes and analysis from comprehensive comparative experiments and ablation studies, conducted to validate the effectiveness and superiority of our proposed method.

\subsection{Training Dataset}
We employ three commonly used datasets from CNR~\cite{Wang-2016-SA}, and individually train \textit{SurfaceFormer} on them:

\noindent \textbf{-- Synthetic dataset.} The dataset contains 21 noise-free synthetic meshes, and they are corrupted by three level of Gaussian noise: 0.1, 0.2 and 0.3 deviations relative to the average edge length, resulting in a total of 63 noisy meshes. These are categorized into: \textit{CAD} (18 meshes with sharp edges), \textit{Smooth} (24 meshes with curved surface and plane), and \textit{Feature} (21 meshes with rich details). This dataset contains 3.0 million faces.

\noindent \textbf{-- Kinect V1 dataset.}  The dataset are captured from four statues with both Microsoft Kinect v1 and a high accuracy scanner. The collection comprises \textit{Big Girl} (24 meshes),  \textit{Cone} (12 meshes), \textit{David} (23 meshes) and \textit{Pyramid} (12 meshes). Meshes captured by Microsoft Kinect v1 serve as the noisy instances, while those from the high-accuracy scanner serve as the ground truth. This dataset contains 2.4 million faces.

\noindent \textbf{-- Kinect V2 dataset.} Analogous in structure to the Kinect V1 dataset but utilizing Microsoft Kinect v2 for scanning, this set includes \textit{Big Girl} (24 meshes),  \textit{Cone} (12 meshes), \textit{David} (24 meshes), \textit{Pyramid} (12 meshes). This dataset contains 0.9 million faces.

\subsection{Testing Dataset}
We conduct experimental on five datasets. The first three, released by~\cite{Wang-2016-SA}, are built using the same methodology as their training counterparts, comprising  Synthetic, Kinect V1, and Kinect V2 datasets. These datasets are equipped with ground truth meshes, enabling both objective and subjective comparison. We utilize the \textit{SurfaceFormer} trained on the corresponding training set to obtain the best results. Additionally, to evaluate the generalization capabilities of the compared methods, we introduce a Real-scanned dataset from ~\cite{zhang2015guided,8012522}, obtained from unknown scanners. Finally, to validate the performance of \textit{SurfaceFormer} in practical applications, we choose four point clouds from different scanned datasets and utilize  the Poisson surface reconstruction~\cite{kazhdan2013screened} to build noisy meshes as the Reconstructed dataset. These two datasets lack ground truth meshes and are subject to subjective evaluation only. The \textit{SurfaceFormer} trained on the Synthetic dataset is applied to denoise these meshes. 

The specifics of each test dataset are as follows:

\noindent \textbf{-- Synthetic dataset}:  \textit{CAD} (42 meshes),  \textit{Smooth} (21 meshes) and \textit{Feature} (24 meshes). 

\noindent \textbf{-- Kinect V1 datasets}:  \textit{Boy} (24 meshes), \textit{Girl} (25 meshes), \textit{Cone} (12 meshes) and \textit{Pyramid} (12 meshes).

\noindent \textbf{-- Kinect V2 datasets}:  \textit{Boy} (24 meshes), \textit{Girl} (24 meshes), \textit{Cone} (12 meshes) and \textit{Pyramid} (12 meshes). 

\noindent \textbf{-- Real-scanned dataset}:  \textit{Angel} and \textit{Eagle}.

\noindent \textbf{-- Reconstructed dataset}: \textit{Bed} from ScanObjectNN~\cite{uy2019revisiting}, a sparse object dataset; \textit{ScrewNut} from Visionair~\cite{visionair}, another object dataset scanned by different device; \textit{Pillar} from sydney urban objects dataset~\cite{de2013unsupervised}, a outdoor LiDAR dataset; \textit{Soldier} from 8iVFB v2~\cite{G-PCC3}, a dense photo-realistic dynamic datasets.


\renewcommand{\arraystretch}{1.2}
\begin{table*}[htbp]

\setlength\tabcolsep{4pt}
\begin{center}
  \centering
    \begin{tabular}{c|c||c|c|c|c|c|c|c||c|c|c|c|c|c|c}
   \hline
    \multicolumn{2}{c||}{Metric} & \multicolumn{7}{c||}{$E_a$ (Degree)}                      & \multicolumn{7}{c}{$E_v (10^{-1})$ } \\
\hline
      \multicolumn{2}{c||}{Method} & \multicolumn{1}{c|}{CNR} & \multicolumn{1}{c|}{NFN} & \multicolumn{1}{c|}{NNT} & \multicolumn{1}{c|}{GCN} & \multicolumn{1}{c|}{GEO} & \multicolumn{1}{c|}{RES} & \multicolumn{1}{c||}{Ours} & \multicolumn{1}{c|}{CNR} & \multicolumn{1}{c|}{NFN} & \multicolumn{1}{c|}{NNT} & \multicolumn{1}{c|}{GCN} & \multicolumn{1}{c|}{GEO} & \multicolumn{1}{c|}{RES} & \multicolumn{1}{c}{Ours} \\
\hline

    \multirow{4}[4]{*}{Synthetic} & \textit{CAD}   & 3.71  & 5.00  & 3.46  & 3.29  & 2.99  &  \blue{2.50}  & \red{2.05}  & 0.52  & 0.79  & 0.60  & 0.58  & 0.55  & \blue{0.46}  & \red{0.34}  \\
          & \textit{Smooth} & 3.08  & 3.22  & 3.06  & 2.62  & 2.73  & \blue{2.43}  & \red{2.18}  & 0.47  & 0.56  & 0.53  & 0.40  & 0.48  & \blue{0.44}  & \red{0.33}  \\
          & \textit{Feature} & 6.34  & 5.74  & 6.71  & 5.08  & 5.14  & \blue{4.72}  & \red{4.32}  & 0.80  & 0.85  & 0.99  & 0.68  & 0.76  & \blue{0.65}  & \red{0.57}  \\
\cline{2-16}          
& Average & 4.28  & 4.77 & 4.26  & 3.62  & 3.52  & \blue{3.10}  & \red{2.71}  & 0.58  & 0.75  & 0.69  & 0.56  & 0.59  & \blue{0.51}  & \red{0.40}  \\

\hline
    \multirow{5}[4]{*}{Kinect V1} & \textit{Boy}   & 8.48  & 9.11  & 8.52  & 8.28  & 7.73  &\blue{6.63}  & \red{6.54}  & 5.47  & 5.64  & 5.42  & 5.37  & 4.58  & \blue{4.31}  & \red{3.71}  \\
          & \textit{Cone}  & 8.80  & 9.47  & 8.87  & 8.73  & 7.39  & \blue{5.80}  & \red{4.29}  & 5.22  & 5.44  & 5.19  & 5.15  & 4.26  & \blue{3.83} & \red{2.67}  \\
          & \textit{Girl}  & 11.45  & 11.94  & 11.43  & 11.13  & 10.73  & \blue{9.32}  & \red{8.39}  & 5.77  & 5.96  & 5.73  & 5.61  & 4.77  & \blue{4.53}  & \red{3.75}  \\
          & \textit{Pyramid} & 8.13  & 9.24  & 8.96  & 9.03  & 7.92  & \blue{6.08}  & \red{4.87}  & 4.95  & 5.20  & 5.10  & 5.09  & 4.13  & \blue{3.80}  & \red{2.88}  \\
\cline{2-16}             & Average & 9.49  & 10.16  & 9.65  & 9.45  & 8.73  & \blue{7.33}  & \red{6.53} & 5.45  & 5.65  & 5.44  & 5.37  & 4.52  & \blue{4.22}  & \red{3.42}  \\
\hline

    \multirow{5}[4]{*}{Kinect V2} & \textit{Boy}   & 7.95  & 8.33  & 7.80  & 7.43  & 7.84  & \blue{6.66}  & \red{5.69}  & 5.27  & 5.26  & 5.19  & 5.10  & 4.21  & \blue{3.74}  & \red{2.65}  \\
          & \textit{Cone}  & 6.34  & 6.62  & 6.45  & 5.86  & 5.98  & \blue{4.74}  & \red{3.47}  & 4.47  & 4.39  & 4.41  & 4.31  & 2.95  & \blue{2.87}  & \red{1.95}  \\
          & \textit{Girl}  & 9.36  & 10.04  & 9.31  & 8.96  & 9.18  & \blue{8.11}  & \red{7.47}  & 4.06  & 4.16  & 4.00 & 3.96  & 3.18  & \blue{3.10}  & \red{2.74}  \\
          & \textit{Pyramid} & 6.76  & 7.04  & 7.04  & 6.44  & 6.69  & \blue{5.18}  & \red{4.14}  & 4.21  & 4.18  & 4.16  & 4.07  & 3.64  & \blue{2.89}  & \red{2.31}  \\
\cline{2-16}           & Average & 7.96  & 8.40  & 7.95  & 7.51  & 7.79  & \blue{6.58}  & \red{5.65}  & 4.56  & 4.57  & 4.49  & 4.42  & 3.56  & \blue{3.24}  & \red{2.51}  \\
\hline
 
  \end{tabular}%
  \label{Ea}%
  \end{center}
  \caption{The objective comparison of $E_a$ and $E_v$, the best and second best results are marked in red and blue, respectively.}
\end{table*}%

\begin{figure*}
\begin{center}
\includegraphics[width=\linewidth]{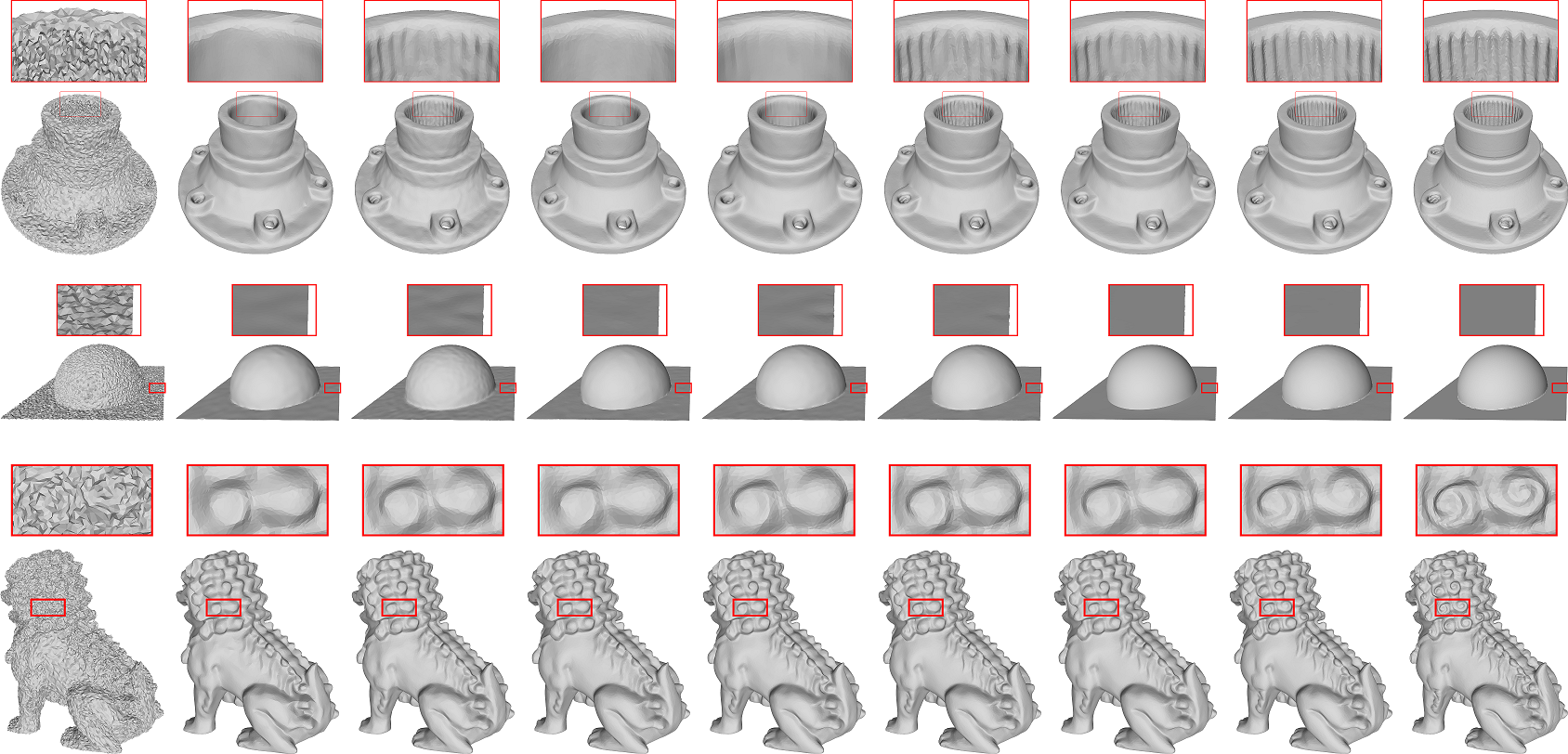}
{(a)\hspace{1.68cm}(b)\hspace{1.68cm}(c)\hspace{1.68cm}(d)\hspace{1.68cm}(e)\hspace{1.68cm}(f)\hspace{1.68cm}(g)\hspace{1.68cm}(h)\hspace{1.68cm}(i)}
\end{center}
   \caption{Denoising results of \textit{CAD} mesh \textit{Carter100K}, \textit{Smooth} mesh \textit{Plane-sphere} and \textit{Feature} mesh \textit{Chinese-Lion} from synthetic dataset~\cite{Wang-2016-SA}. (a) the noisy meshes; (b) CNR~\cite{Wang-2016-SA}; (c) NFN~\cite{li2020normalf}; (d) NNT~\cite{zhao2021normalnet}; (e) GCN~\cite{DBLP:journals/corr/abs-2108-05128}; (f) GEO~\cite{zhang2022geobi}; (g) RES~\cite{zhou2024resgem}; (h) Ours; (i) the ground truth. 
}
\label{r1}
\vspace{-0.3cm}
\end{figure*}

\subsection{Comparison Study}
We evaluate the performance of \textit{SurfaceFormer} against six state-of-the-art methods across all the test datasets. The compared methods can be divided into two categories based on their geometrical representation:

\noindent \textbf{-- Methods using regular representation.} This group includes Cascaded normal regression (CNR)~\cite{Wang-2016-SA},  Normalf-net (NFN)~\cite{li2020normalf} and NormalNet (NNT)~\cite{zhao2021normalnet}. These methods transform noisy meshes into regular representations and leverage CNN or MLP-based networks. 

\noindent\textbf{-- Methods using graph-based representation.} This group includes GCN-Denoiser (GCN)~\cite{DBLP:journals/corr/abs-2108-05128}, GeoBi (GEO)~\cite{zhang2022geobi} and ResGEM (RES)~\cite{zhou2024resgem}. These methods utilize graph-based representations to maintain the irregular structure of meshes and leverage GNN-based networks. 

For CNR~\cite{Wang-2016-SA} and GCN~\cite{DBLP:journals/corr/abs-2108-05128}, the authors kindly provided their code and pre-trained models of the three training datasets, allowing for direct application to the respective test datasets. For GEO~\cite{zhang2022geobi}, the authors provided the code and we strictly follow their paper to retrain the models on the training datasets. For NNT~\cite{zhao2021normalnet}, NFN~\cite{li2020normalf} and RES~\cite{zhou2024resgem}, the authors kindly provided us their denoised meshes for comparison.

\subsection{Parameters Setting}
For the iteration number in \textit{Vertex Refinement}, we set $N_v=60$. For the parameter in \textit{Mesh Denoising}, we set the feature dimension $D=512$, the number of Transformer encoder $L=12$, the number of heads $N_h=12$.  Following Vit~\cite{dosovitskiy2020image}, the MLP size is set to $4 \times D$. The choice of these parameters will be further discussed in \textit{Ablation Study}.

For the parameters in \textit{Patch Generation}, we set the face number of each patch $T_f=240$, further increasing $T_f$ will exceed the memory limitation of GPUs. For the parameters in \textit{Surface Representation}, we set the precision of sampling $p_s=8$, which aligns with the setting in prior work~\cite{zhao2021normalnet} for precise sampling.
We set $T_s=1.25\times p_s$ to ensure the LSD contains 1-ring faces, which maintains adjacency relationships. The direction of normalization $\n_{t}$ can be arbitrary and we set $\n_{t}=\left(1,0,0\right)$.

\begin{figure*}
\begin{center}
\includegraphics[width=\linewidth]{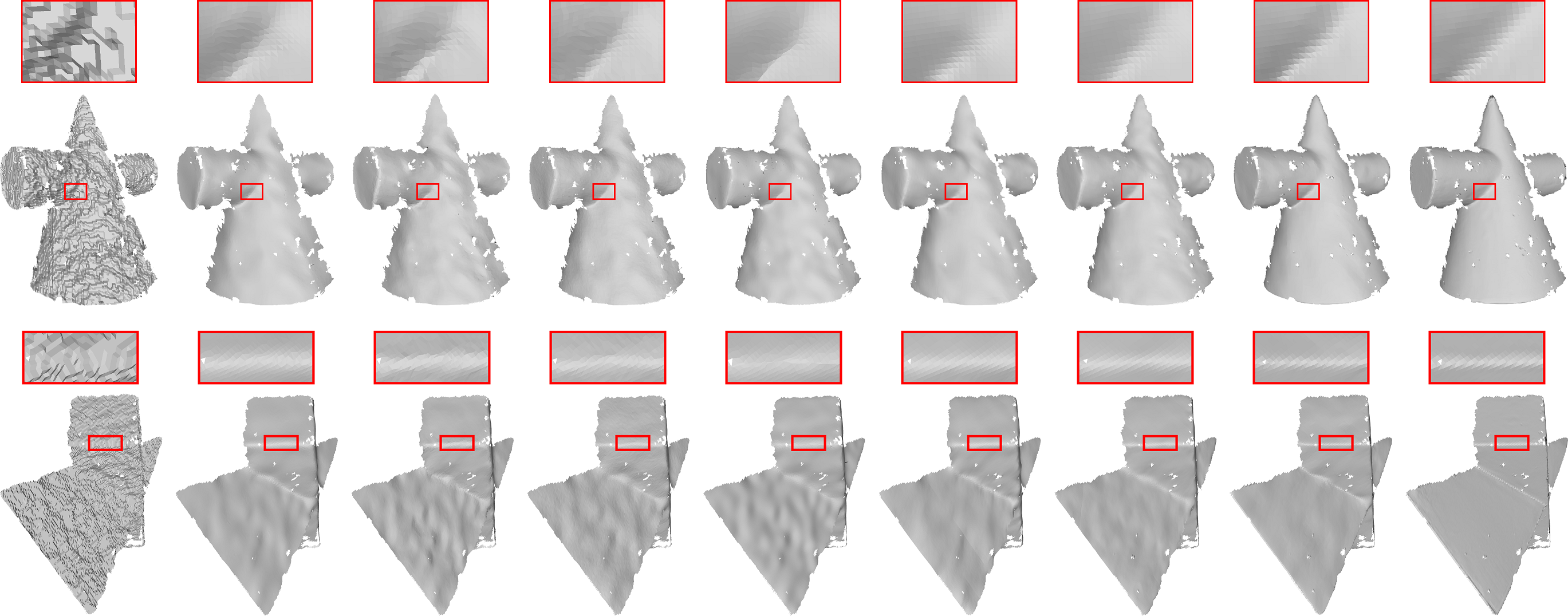}
{(a)\hspace{1.68cm}(b)\hspace{1.68cm}(c)\hspace{1.68cm}(d)\hspace{1.68cm}(e)\hspace{1.68cm}(f)\hspace{1.68cm}(g)\hspace{1.68cm}(h)\hspace{1.68cm}(i)}
\end{center}
   \caption{Denoising results of \textit{Cone\_16} and \textit{Pyramid\_04} from Kinect V1 dataset~\cite{Wang-2016-SA}.  (a) the noisy meshes; (b) CNR~\cite{Wang-2016-SA}; (c) NFN~\cite{li2020normalf}; (d) NNT~\cite{zhao2021normalnet}; (e) GCN~\cite{DBLP:journals/corr/abs-2108-05128}; (f) GEO~\cite{zhang2022geobi}; (g) RES~\cite{zhou2024resgem}; (h) Ours; (i) the ground truth. 
}
\label{r2}
\vspace{-0.3cm}
\end{figure*}

\subsection{Objective Performance Comparison}
\textbf{-- Evaluation Metrics.}  To assess our results and quantitatively contrast our method with the state-of-the-art methods, we employ two standard metrics, $E_a$ and $E_v$, from GCN~\cite{DBLP:journals/corr/abs-2108-05128}, which quantify errors in face normals and vertex positions, respectively. Specifically, ${E_a}$ is the average normal angular difference between the denoised normals and ground truth:
\begin{equation}
{E_a} = \frac{1}{{{S_f}}}\sum\limits_{i \in [1,S_f] } {acos \left( {\widehat \n_i \cdot \n_i^*} \right)} 
\end{equation}
where $S_f$ is total number of faces. 

${E_v}$ is the normalized one-sided average Hausdorff distance from denoised vertices to the ground truth:
\begin{equation}
{E_v} = \frac{1}{{{S_v}{L_d}}}\sum\limits_{\widehat \v_i \in \widehat \x} {\mathop {\min }\limits_{\v_j^* \in {\x^*}} \left\| {\widehat \v_i - \v_j^*} \right\|} 
\end{equation}
where $S_v$ is total number of vertices, $L_d$ is the average edge length of the ground truth mesh. Lower values of both metrics indicate higher quality.

\noindent\textbf{-- Objective Comparison.} We present the comparison results of $E_a$ and $E_v$ in Table 1. For $E_a$, our method ranks first across all datasets, exhibiting average improvements of 12.6\%, 10.9\%, and 14.0\% over the second-ranked method for each dataset, respectively. For $E_v$, our method also ranks first and achieves similar improvements on different dataset: 20.9\%, 19.7\%, and 22.5\%, respectively. This demonstrates the effectiveness and robustness of our method in handling diverse noise types. Furthermore, due to the proposed LSD's efficient preservation of local feature structures, our method demonstrates superior performance on meshes with significant features, such as the \textit{CAD} category in the Synthetic dataset ($E_a$: 18.1\%,  $E_v$: 25.5\%) and \textit{Cone} category in the Kinect V1 dataset ($E_a$: 26.1\%,  $E_v$: 30.1\%). This will be further demonstrated in the subsequent subjective comparison.


\noindent\textbf{-- Time Efficiency.} We evaluated our method's efficiency on the same server used for training. Statistically, generating LSD operates at a rate of approximately 3980 LSDs per second on a single CPU core, and denoising operates at a rate of 144 patches per second on two GPUs. For better understanding, we take the Synthetic model \textit{Child} as an example, which comprises 100K faces and is divided into 1208 patches. Our method requires only 25.12 seconds for LSD generation and 8.37 seconds for denoising. This performance is comparable with GCN~\cite{DBLP:journals/corr/abs-2108-05128} , which takes 30 seconds to denoise \textit{Child}. Furthermore, both LSD generation and denoising can be parallelized across multiple CPU cores and GPUs, respectively, potentially further reducing computation times.

\begin{figure*}
\begin{center}
\includegraphics[width=\linewidth]{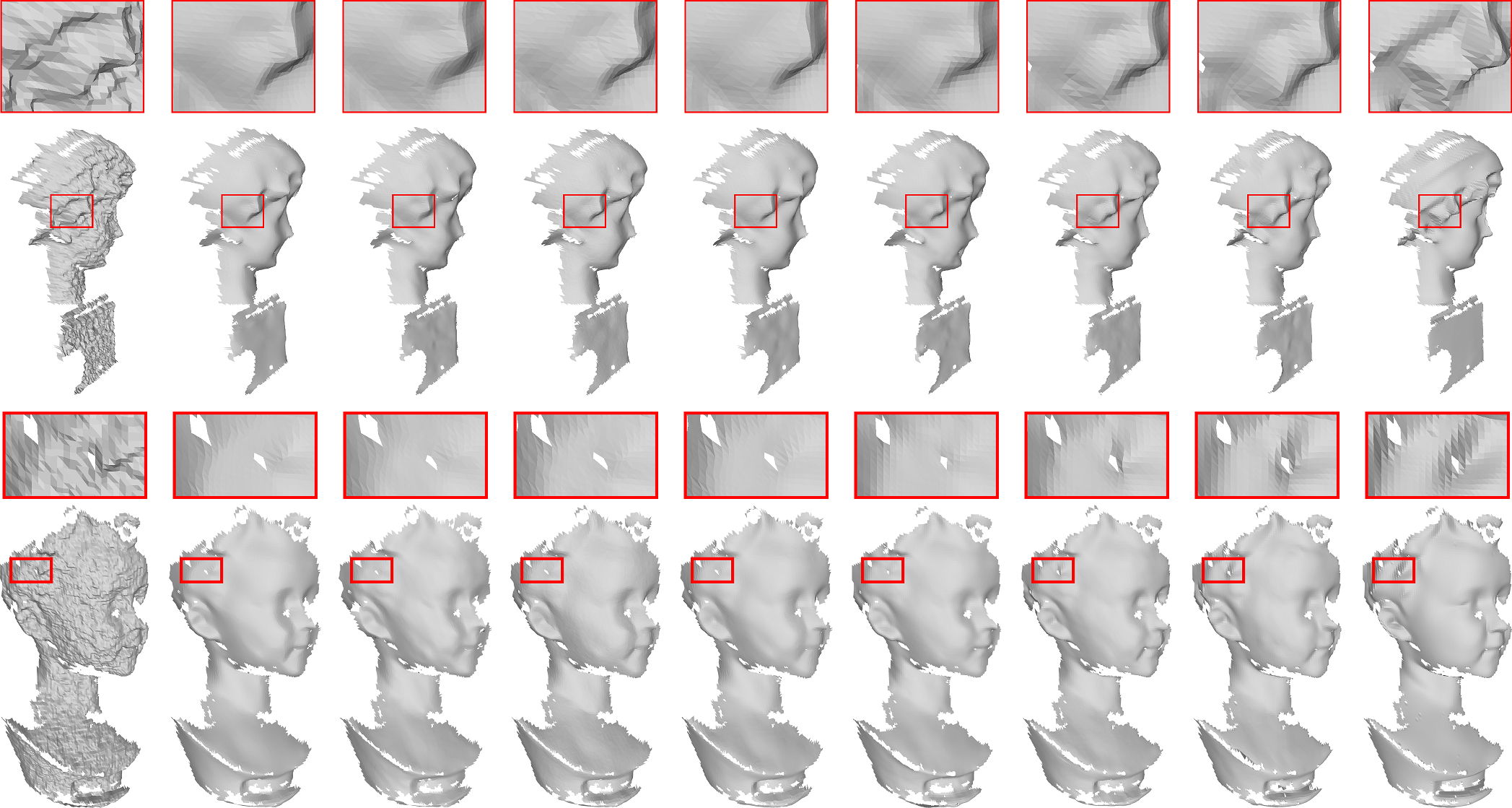}
{(a)\hspace{1.68cm}(b)\hspace{1.68cm}(c)\hspace{1.68cm}(d)\hspace{1.68cm}(e)\hspace{1.68cm}(f)\hspace{1.68cm}(g)\hspace{1.68cm}(h)\hspace{1.68cm}(i)}
\end{center}
   \caption{Denoising results of \textit{Girl\_04} and \textit{Boy\_05} from Kinect V2 dataset~\cite{Wang-2016-SA}. (a) the noisy meshes; (b) CNR~\cite{Wang-2016-SA}; (c) NFN~\cite{li2020normalf}; (d) NNT~\cite{zhao2021normalnet}; (e) GCN~\cite{DBLP:journals/corr/abs-2108-05128}; (f) GEO~\cite{zhang2022geobi}; (g) RES~\cite{zhou2024resgem}; (h) Ours; (i) the ground truth.   }
\label{r3}
\vspace{-0.3cm}
\end{figure*}

\subsection{Subjective Performance Comparison}
\noindent \textbf{-- Synthetic Dataset.} In Fig.~\ref{r1}, we present the visual comparison of \textit{CAD} mesh \textit{Carter100K}, \textit{Smooth} mesh \textit{Plane-sphere} and \textit{Feature} mesh \textit{Chinese-lion}. Our method outperforms others in both feature recovery and noise removal. Specifically, for \textit{Carter100K} and \textit{Chinese-lion},  the proposed LSD effectively retains rich mesh attributes, enabling precise differentiation of small-scale features from noise and leading to superior feature recovery. In contrast, other methods generally result in over-smoothing.  For \textit{Plane-sphere}, due to our method's capability to aggregate global information, it successfully avoid mistaking local noise for feature structures, whereas other methods leave artifacts.

\noindent \textbf{-- Kincet Dataset.} In Fig.~\ref{r2}, we
illustrate the visual comparison of \textit{Cone\_16} and \textit{Pyramid\_04} from Kincet V1 Dataset. These meshes contains large-scale features with distributions distinct from synthetic meshes. Our method adeptly learns noise patterns to provide the best feature recovery for both meshes, while GCN~\cite{DBLP:journals/corr/abs-2108-05128}, GEO~\cite{zhang2022geobi} and RES~\cite{zhou2024resgem} output over-smooth results, CNR~\cite{Wang-2016-SA}, NFN~\cite{li2020normalf} and NNT~\cite{zhao2021normalnet} struggle with noise removal, respectively. In Fig.~\ref{r3}, we show the visual comparison of \textit{Girl\_04} and \textit{Boy\_05} from Kincet V2 Dataset. These meshes contains rich detail and small-scale features. In \textit{Girl\_04}, only our method and RES~\cite{zhou2024resgem} are able to preserve feature structures, but RES~\cite{zhou2024resgem} fails to remove noise. In \textit{Boy\_05}, our method can preserves the feature while all the other method introduce over-smoothing in the zoom-in region.

\begin{figure*}
\begin{center}
\includegraphics[width=\linewidth]{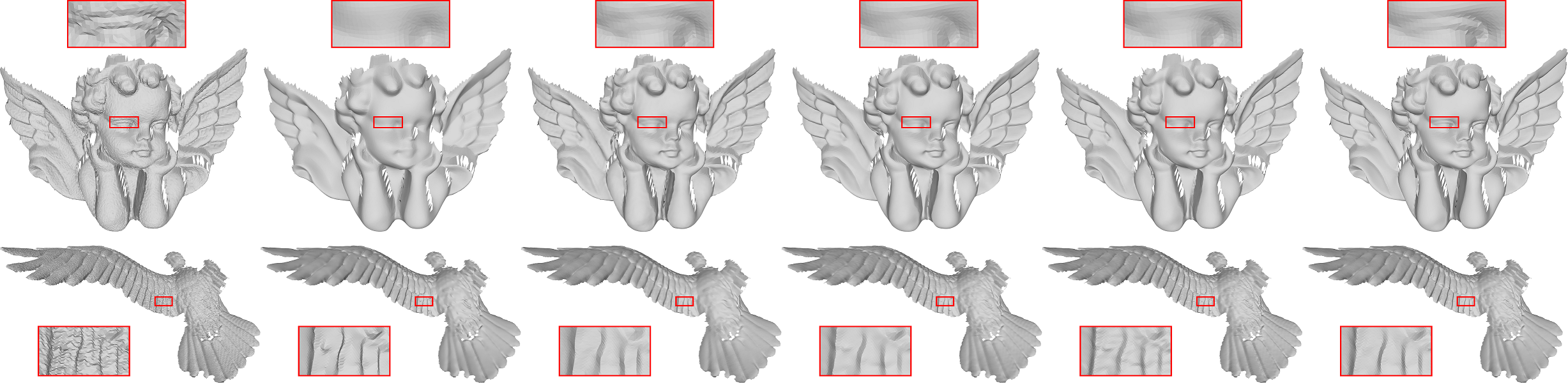}
{(a)\hspace{2.65cm}(b)\hspace{2.65cm}(c)\hspace{2.65cm}(d)\hspace{2.65cm}(e)\hspace{2.65cm}(f)}
\end{center}
   \caption{Denoising results of \textit{Angel} and \textit{Eagle} from Real-scanned dataset~\cite{zhang2015guided,8012522}. (a) the noisy meshes; (b) CNR~\cite{Wang-2016-SA}; (c) NNT~\cite{zhao2021normalnet}; (d) GCN~\cite{DBLP:journals/corr/abs-2108-05128}; (e) GEO~\cite{zhang2022geobi}; (f) Ours.
}
\label{r4}
\end{figure*}

\noindent \textbf{-- Real-Scanned Dataset.}  To further assess our method's robustness, in Fig.~\ref{r4}, we present the visual comparison of \textit{Angel} and \textit{Eagle}. Since that the authors of NFN~\cite{li2020normalf} and RES~\cite{zhou2024resgem} do not provide the results of these meshes, they are not included in this comparison. In \textit{Angel}, 
CNR~\cite{Wang-2016-SA} and GEO~\cite{zhang2022geobi} fail to catch the feature structures, resulting in over-smoothing outputs. NNT~\cite{zhao2021normalnet} and GCN~\cite{DBLP:journals/corr/abs-2108-05128} successfully perceive feature structures but cannot accurately restore the narrow edges. In contrast, our method produces the best feature recover results. In \textit{Eagle}, CNR~\cite{Wang-2016-SA}, GCN~\cite{DBLP:journals/corr/abs-2108-05128} and GEO~\cite{zhang2022geobi} can not remove noise, while NNT~\cite{zhao2021normalnet} can remove noise but introduces pseudo-feature. Our method accurately removes noise while restoring feature structures.

\begin{figure*}
\begin{center}
\includegraphics[width=\linewidth]{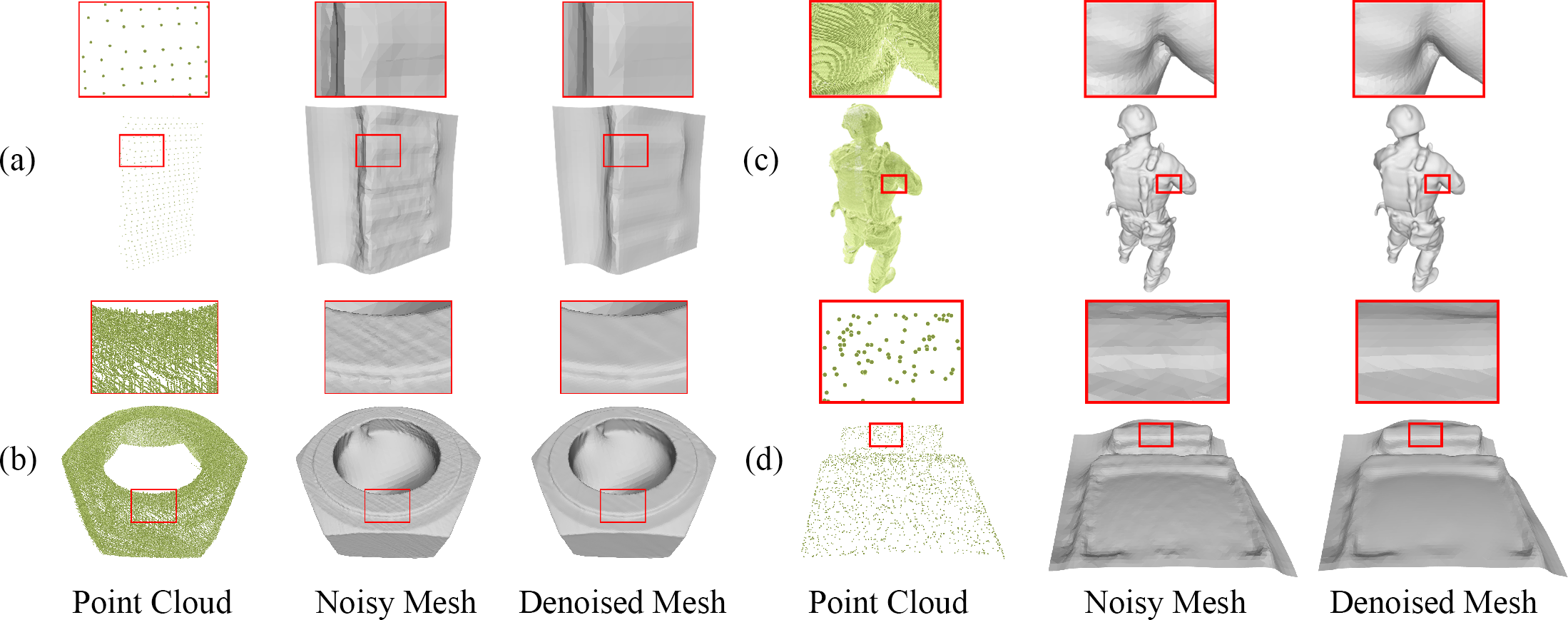}
\end{center}

   \caption{Denoising results of Reconstructed dataset. (a) \textit{Pillar} from sydney urban objects dataset~\cite{de2013unsupervised}; (b) \textit{ScrewNut} from Visionair~\cite{visionair}; (c)\textit{Soldier} from 8iVFB v2~\cite{G-PCC3}; (d) \textit{Bed} from ScanObjectNN~\cite{uy2019revisiting}.
}

\label{r5}
\end{figure*}

\noindent \textbf{-- Reconstructed Dataset.}  In real-world applications, a major resource for meshes stems from the reconstruction of point clouds. However, since acquired point clouds are typically sparse, the resulting meshes often suffer from reconstruction noise. In view of this issue, we evaluate the performance of \textit{SurfaceFormer} on mitigating reconstruction noise. We present the denoising results in Fig.~\ref{r5}. Our method effectively manages different noise distributions from various scanners and achieves satisfactory feature recovery results in \textit{ScrewNut} and \textit{Pillar}. Moreover, our approach produces smooth surfaces in \textit{Bed} and \textit{Soldier}, avoiding the introduction of pseudo-features. This demonstrates that our algorithm accurately distinguishes features from noise in practical applications and achieves effective denoising.

These experiments underscore our algorithm's adeptness at restoring features across meshes produced by various scanning technologies.

\begin{table}
\begin{center}
\setlength\tabcolsep{3pt}
\begin{tabular}{c|c|c||c|c|c|c}
\hline
Linear & Shape & Spatial &  \textit{CAD} & \textit{Smooth} & \textit{Feature} &Avg. \\
\hline
$\checkmark$ & $\times$& $\times$ &	2.11	&   2.33&	4.57& 2.84 (104\%)\\
\hline
$\checkmark$ &$\checkmark$ & $\times$ &	2.11 &	2.24&	4.49& 	2.80 (102\%)\\
\hline
$\checkmark$ &$\checkmark$ & $\checkmark$ &	2.08 & 2.18 & 4.35& 2.73 (100\%)\\
\hline

\end{tabular}
\end{center}

\caption{Ablation study on the Network structure.}
\label{ab0}
\end{table}




\begin{table}
\begin{center}
\setlength\tabcolsep{3pt}
\begin{tabular}{c||c|c|c|c|c}
\hline
Network &  \textit{CAD} & \textit{Smooth} & \textit{Feature} & Average. & Denoising Speed \\
\hline
\textit{Small} & 2.24 &  2.34  &	4.60 & 2.91 (107\%)& 161 LSD/s (112\%)\\
\hline
\textit{Middle} & 2.08 & 2.18 & 4.35& 2.73 (100\%)& 144 LSD/s (100\%)\\
\hline
\textit{Large}  & 2.01 & 2.07 & 4.31  & 2.66 (97\%)& 99 LSD/s (69\%)\\
\hline

\end{tabular}
\end{center}

\caption{Ablation study on the network parameters.}
\label{ab2}
\vspace{-0.6cm}
\end{table}







\subsection{Ablation Study}

\begin{figure*}
\begin{center}
\includegraphics[width=\linewidth]{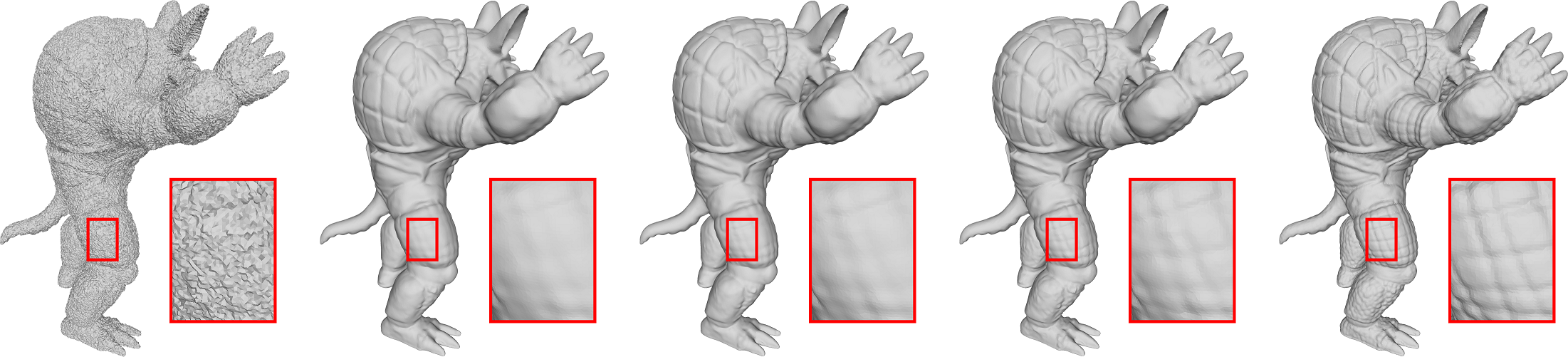}
{(a)\hspace{3.3cm}(b)\hspace{3.3cm}(c)\hspace{3.3cm}(d)\hspace{3.3cm}(e)}
\end{center}
   \caption{Denoising results of \textit{Feature} mesh \textit{Armadillo} from synthetic dataset~\cite{Wang-2016-SA}. (a) the noisy meshes; (b) Linear embedding; (c) Linear embedding + shape embedding; (d) Full pipeline; (e) the ground truth. 
}
\label{abf1}
\end{figure*}

\noindent\textbf{-- About the network structure.}
To investigate the importance of various components of \textit{SurfaceFormer}, we provide ablation study to verify the effect of the Shape Embedding and Spatial Encoder on the Synthetic dataset. From Table 2, it can be observed that both the introduce of Shape Embedding and Spatial Encoder lead to a performance improvement on all categories, providing evidence of the effectiveness of our designed layers. Moreover, the introduction of spatial and edge information enhances the performance of distinguishing features: As shown in Fig.~\ref{abf1}, the zoom-in region contains small-scale feature, and the performance of feature recovering gradually improves with the introduction of these components.

\begin{figure*}
\begin{center}
\includegraphics[width=\linewidth]{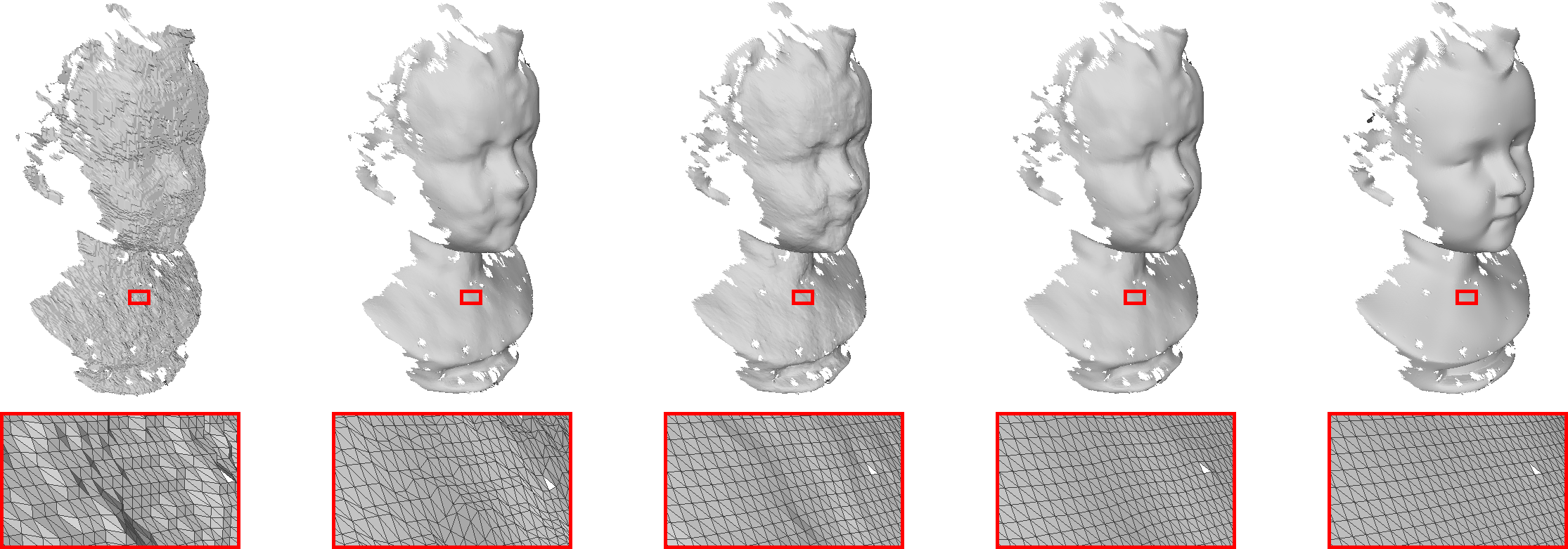}
{(a)\hspace{3.4cm}(b)\hspace{3.4cm}(c)\hspace{3.4cm}(d)\hspace{3.4cm}(e)}
\end{center}
   \caption{Denoising results of mesh \textit{Boy\_01} from Kinect V1 dataset. (a) the noisy meshes; (b) denoising result in normal domain; (c) denoising result in vertex domain; (d) denoising result in multi domain; (e) the ground truth.}

\label{abf2}
\end{figure*}

\noindent\textbf{-- About the multi-domain denoising.}
Mesh denoising can be achieved in normal domain or vertex domain alone. However, integrating multi-domain information yields better results. To verify this, we apply \textit{SurfaceFormer} to the \textit{Boy\_01} mesh from the Kinect V1 dataset, and obtain the denoised mesh through three methods: (1) normal domain denoising: employing the vertex updating method~\cite{zhang2015guided} to get denoised vertex according to the denoised normals; (2) vertex domain denoising: directly using the denoised vertex coordinates~\cite{fleishman2003bilateral}; (3) multi-domain denoising: aligning the vertex coordinates to the normals as mentioned in \textit{Vertex refinement}. The denoising results are shown in Fig.~\ref{abf2}.  Specifically, method 1 can output smooth surface, but the distribution of vertices is corrupted by artificial effects. This is because the updating method~\cite{zhang2015guided} cannot catch the underlying distribution patterns of vertices. Method 2 can output vertices with a distribution similar to the ground truth. However, vertex coordinates exhibit less consistency compared to normals, resulting in rough surface. Method 3 integrates the consistency of normal and the distribution vertex, leads to a satisfactory denoising effect.

\begin{figure}
\begin{center}
\includegraphics[width=\linewidth]{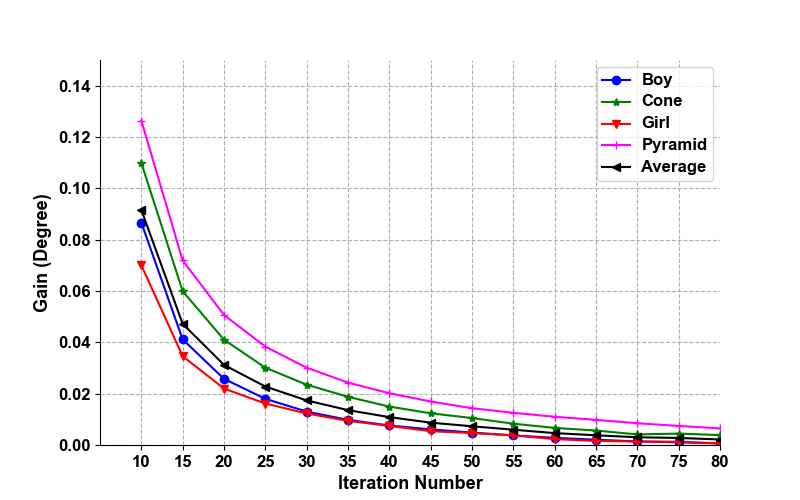}
\end{center}
   \caption{The gain of $E_a$ for every 5 iterations on Kinect V2 dataset.
}
\label{abf3}
\end{figure}


\noindent\textbf{-- About the choice of network parameter.}  
We conducted ablation study about three network parameters: the size of latent vector $D$, the number of Transformer encoders $T$ and the number of attention heads $N_h$. To find the best combination of these parameters, we train three levels of networks with different parameters: \textit{Small}: $D=256$, $T=8$, $N_h=8$, \textit{Middle} (Proposed): $D=512$, $T=12$, $N_h=12$ and \textit{Large}: $D=1024$, $T=16$, $N_h=16$, then test them on the Synthetic dataset. As shown in Table 3, when the network parameters increase from \textit{Small} to \textit{Middle}, the denoising performance improves by 7\% while the processing speed decreases by 12\%. However, when the network parameters increase from \textit{Middle} to \textit{Large}, the performance only improves by 3\% while the processing speed decreases by 31\%. Considering the balance between speed and performance, we choose the \textit{Middle} configuration ($D=512$, $T=12$, $N_h=12$).

\noindent\textbf{-- About the choice of $N_v$.} 
We provide ablation study about the iteration number $N_v$ in \textit{Vertex Refinement}. To determine the optimal choice of $N_v$, we perform \textit{Vertex Refinement} on the Kinect V2 dataset, saving the denoised results after every 5 iterations, and then calculate the gain of $E_a$ between adjacent results. As shown in Fig.~\ref{abf3}, the gain of each iteration decreases gradually as the iteration number increases. When the iteration number reaches 60, further increasing the iteration number would not result in a significant gain but only increase processing time. Therefore, we choose $N_v=60$.


\section{Conclusion }
\label{sec:conclusion}

Existing deep learning-based mesh denoising methods are constrained by single-modal geometric representations and network structures, limiting their capacity to capture the intricate attributes of meshes and accomplish effective global feature aggregation. In this paper, we introduced a novel and potent mesh denoising method, \textit{SurfaceFormer}. Our first contribution is a novel multi-model geometric representation called \textit{Local Surface Descriptor} (LSD), which encodes local geometric details and spatial information into a sequence of 2D matrices and a point cloud, respectively. Next, we propose a dual-stream structure consisting of a Geometric Encoder branch and a Spatial Encoder branch, which jointly explore multimodal information for mesh denoising. A subsequent Denoising Transformer module receives the multimodal information and achieves efficient global feature aggregation through self-attention operators. Extensive experimental results demonstrate that our proposed method achieves superior performance in both objective and subjective evaluations compared to state-of-the-art mesh denoising techniques.

\bibliography{egbib.bib}

\bibliographystyle{IEEEtran}

\end{document}